\documentclass[sn-basic]{sn-jnl}% Basic Springer Nature Reference Style/Chemistry Reference Style
\usepackage{diagbox}
\usepackage{pifont}
\usepackage{graphicx}%
\usepackage{multirow}%
\usepackage{amsmath,amssymb,amsfonts}%
\usepackage{amsthm}%
\usepackage{mathrsfs}%
\usepackage[title]{appendix}%
\usepackage{xcolor}%
\usepackage{textcomp}%
\usepackage{manyfoot}%
\usepackage{booktabs}%
\usepackage{algorithm}%
\usepackage{algorithmicx}%
\usepackage{algpseudocode}%
\usepackage{listings}%
\usepackage{soul}

\def \ie {\emph{i.e.}~}
\def \eg {\emph{e.g.}~}
\def \etc {\emph{etc.}~}
\def \etal {\emph{et al.}~}

\begin{document}

\title[Article Title]{Bootstrapping Vision-language Models for Frequency-centric Self-supervised Remote Physiological Measurement}


\author[1]{\fnm{Zijie} \sur{Yue}}
% \email{zijie@tongji.edu.cn}

\author*[1,2]{\fnm{Miaojing} \sur{Shi}}\email{mshi@tongji.edu.cn}

\author[1]{\fnm{Hanli} \sur{Wang}}

\author[3]{\fnm{Shuai} \sur{Ding}}

\author[1]{\fnm{Qijun} \sur{Chen}}

\author[3]{\fnm{Shanlin} \sur{Yang}}

\affil[1]{College of Electronic and Information Engineering, Tongji University, China}

\affil[2]{Shanghai Institute of Intelligent Science and Technology, Tongji University, China}

\affil[3]{School of Management, Hefei University of Technology, China.}




\abstract{Facial video-based remote physiological measurement is a promising research area for detecting human vital signs (\eg, heart rate, respiration frequency) in a non-contact way. Conventional approaches are mostly supervised learning, requiring extensive collections of facial videos and synchronously recorded photoplethysmography (PPG) signals. To tackle it, self-supervised learning has recently gained attentions; due to the lack of ground truth PPG signals, its performance is however limited. In this paper, we propose a novel {frequency-centric} self-supervised framework that successfully integrates the popular vision-language models (VLMs) into the remote physiological measurement task. Given a facial video, we first augment its positive and negative video samples with varying rPPG signal frequencies. Next, we introduce a frequency-oriented vision-text pair generation method by carefully creating contrastive spatio-temporal maps from positive and negative samples and designing proper text prompts to describe their relative ratios of signal frequencies. A pre-trained VLM is employed to extract features for these formed vision-text pairs and estimate rPPG signals thereafter. We develop a series of {frequency-related} generative and contrastive learning mechanisms to optimize the VLM, including the text-guided visual reconstruction task, the vision-text contrastive learning task, and the frequency contrastive and ranking task. Overall, our method for the first time adapts VLMs to digest and align the frequency-related knowledge in vision and text modalities. Extensive experiments on four benchmark datasets demonstrate that it significantly outperforms state of the art self-supervised methods.}

\maketitle
\section{Introduction}\label{sec1}
Heart rate (HR), heart rate variability (HRV) and respiration frequency (RF) serve as important indicators of the physical health for individuals. Two kinds of skin-contact medical sensors, \ie, electrocardiography (ECG) and photoplethysmography (PPG) sensors, are widely adopted to measure these physiological signals. However, the usage of skin-contact sensors can inevitably cause inconvenience and discomfort to participants, particularly in cases of long-term or daily health monitoring \cite{mcduff_camera_2023,niu_rhythmnet_2020,verkruysse_remote_2008}. Moreover, newborns and burn patients cannot use them \cite{chen_video-based_2019,lu_dual-gan_2021}. To cope with these limitations, facial video-based remote physiological measurement technique has emerged and garnered increasing research attentions in recent years \cite{yu_physformer_2022,du_dual-bridging_2023,lu_neuron_2023}. It exhibits promising performance in various applications such as face anti-spoofing \cite{liu_learning_2018}, atrial fibrillation screening \cite{yan_high-throughput_2020}, and vital signs monitoring for ICU patients \cite{jorge_non-contact_2022}.

The core principle of facial video-based remote physiological measurement lies in the remote photoplethysmography (rPPG) technique \cite{wang_algorithmic_2017}, \ie, the optical absorption of skins changes periodically in sync with the cardiac cycle. Ideally, the temporal variation of the skin color reflect the periodic rPPG signal, enabling further measurement of physiological signals such as HR, HRV and RF. However, rPPG signal is inherently intertwined with assorted non-periodic noises arising from illumination variations, subject movements and camera quantization noise \cite{lu_dual-gan_2021,yue_multimodal_2022}. In early studies, researchers developed blind signal separation \cite{wang_algorithmic_2017} and color space transformation \cite{de_haan_robust_2013} approaches to disentangle rPPG signals from noises. However, these methods require manually defining the frequency bandpasses of interest to filter out unexpected noises. They also rely on assumptions about the scene background to diminish the effect of illumination variations on skin tones. As a result, these techniques are primarily suited for the well-controlled laboratory environment and often show severe performance degradation in the real-world environment.

In recent years, many deep learning-based approaches have been proposed, showcasing state of the art performance \cite{yu_physformer_2022,lu_dual-gan_2021,qiu_evm-cnn_2019,vedaldi_meta-rppg_2020,yu_remote_2019,li_learning_2023,niu_video-based_2020}. They leverage diverse vision encoders to extract the visual features from facial videos for rPPG estimation. Nevertheless, the process of collecting facial videos and corresponding ground truth signals (\ie, ECG or PPG signals) for training is still costly, as it requires subjects to wear skin-contact sensors to record videos and signals synchronously. 
% Additionally, ECG or PPG signals often fail to well represent the rPPG signals, as they suffer from a phase delay problem caused by different pulse transit time \cite{sun_contrast-phys_2024}. 
To overcome this challenge, a number of self-supervised approaches have been introduced. They train the models based on unlabeled facial videos using contrastive learning \cite{yang_simper_2023,yue_facial_2023,gideon_way_2021,sun_contrast-phys_2024} or masked auto-encoding \cite{liu_rppg-mae_2024} mechanisms. However, their performance is inevitably inferior to the supervised ones.

% These methods also implicitly solve the 
% \ZJ{Additionally, the estimated rPPG signals and ground truth signals describe pulsations in different parts of the body, so they exhibit a phase delay problem due to different. Consequently, the misalignment between videos and ground truth signals frequently occur during data collection stage . 

% \begin{figure}[!t]
%   \centerline{\includegraphics[width=3.6in]{figs/fig1a.png}}
%   %\vspace{-0.1in}
%   \caption{Vision-Language Models (VLMs) are pre-trained on diverse datasets comprising image-text and video-text pairs that detail the variations of object attributes. This enhances the encoders' ability to comprehend the complex dynamics involved in various scenarios.}
%   \label{fig1}
%   % \vspace{-0.1in}
%   \end{figure}

% \begin{figure}[!t]
%   \centerline{\includegraphics[width=3in]{figs/fig1b.png}}
%   %\vspace{-0.1in}
%   \caption{Based on multimodal contrastive learning applied among the generated vision-text pairs and unimodal contrastive learning conducted between different video samples, we enhance the vision encoder to extract periodic pulsation information for rPPG estimation.}
%   \label{fig10}
%   % \vspace{-0.1in}
%   \end{figure}

\begin{figure*}[!t]
  \centerline{\includegraphics[width=5.7in]{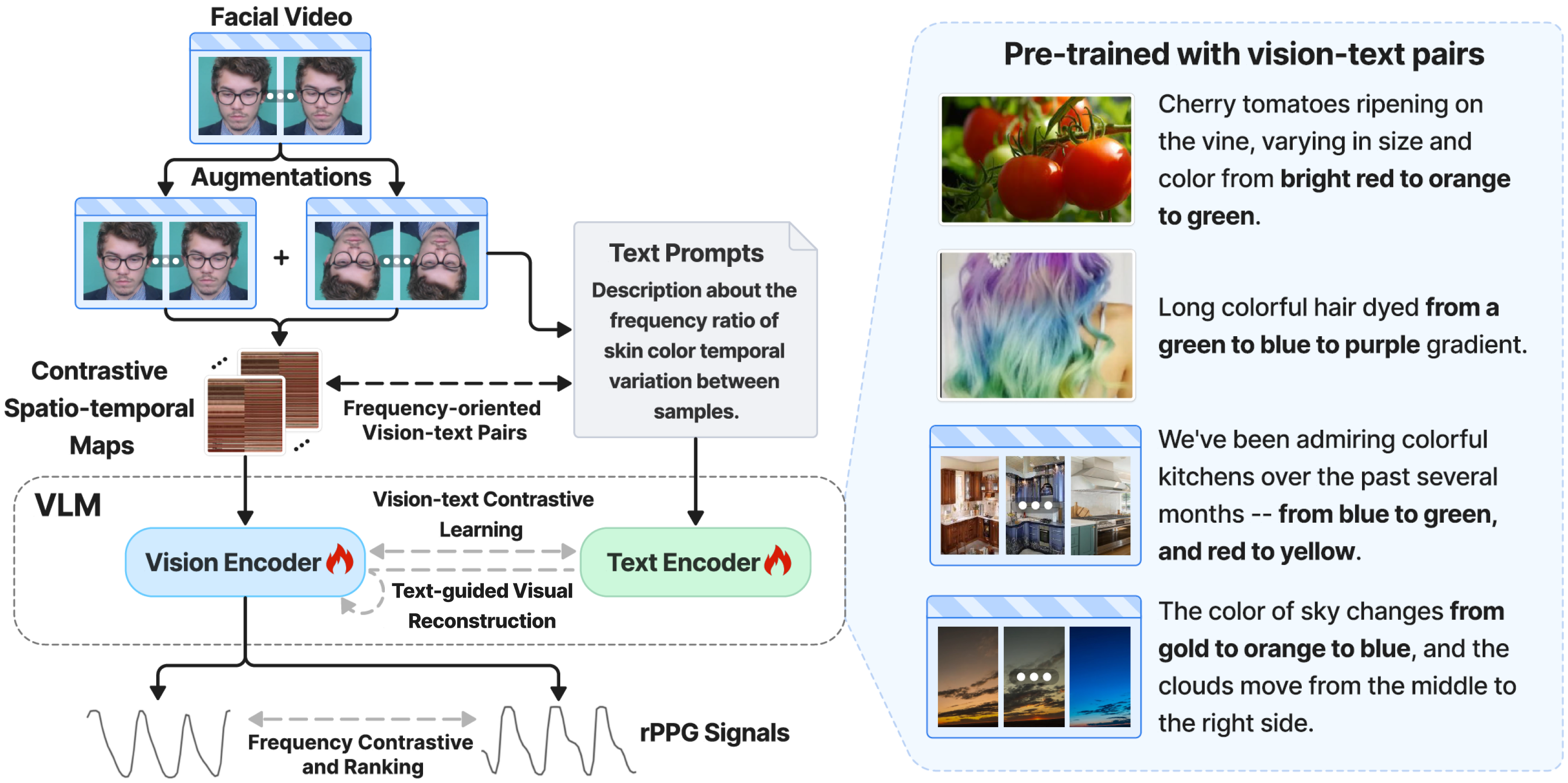}}
  %\vspace{-0.1in}
  \caption{Existing vision-language models (VLMs) are pre-trained on diverse vision-text pairs, including those that describe the temporal variations of certain object attributes. We for the first time adapt VLMs with the ability to digest the frequency-related knowledge of skin color temporal variation in vision and text modalities for self-supervised remote physiological measurement.} 
  % Initially, we augment the given video into multiple samples with varying rPPG signal frequencies. Next, we generate frequency-oriented vision-text pairs by creating contrastive visual maps from them and designing proper text prompts to describe their relative ratios of signal frequencies. Subsequently, we fine-tune the pre-trained encoders of VLMs with these generated pairs for accurate rPPG estimation via a series of generative and contrastive learning mechanisms.
  \label{fig1}
  % \vspace{-0.1in}
  \end{figure*}

Recently, vision-language models (VLMs), such as contrastive language–image pre-training (CLIP) \cite{radford_learning_2021}, video and language understanding (VindLU) \cite{cheng_vindlu_2023}, have drawn increasing research attentions due to their remarkable ability in many vision and language tasks \cite{zhou2025openpsg,zang2024contextual,shi2024llmformer}. Particularly, in video processing, given vision-text pairs, visual and textual embeddings are extracted by vision and text encoders, respectively; and are then aligned via vision-text contrastive learning.
% They apply the multimodal contrastive learning to align the text with vision modality in the embedding space.
As shown in Fig.~\ref{fig1},  the text prompts can be versatile, \ie "the color of sky changes from gold to orange to blue", covering rich temporal details. This type of text cues thus facilitates the  %information from these texts provides additional cues to enhance the
model understanding of temporal visual patterns in the videos.
%for the vision encoder.} 
Therefore, such VLMs have showed special strength in temporal tasks like action recognition \cite{wang_all_2023}, video summarization \cite{pramanick_egovlpv2_2023} and video object segmentation \cite{yuan2024losh}. Inspired by their successes, we aim to leverage the powerful temporal feature modeling ability of pre-trained VLMs to 
% \st{extract periodic pulsation information} 
capture the periodic variation of skin color in facial videos.
% Considering that the frequency of such temporal variation is the key attribute for rPPG estimation.} 
As far as we have explored, VLMs have not been used to capture the frequency-related attributes in images/videos so far. %Inspired by \cite{zhang_can_2022,jiang_clip-count_2023,chatterjee_robustness_2024,zhou_text_2024}, 
Intuitively, we posit two potential solutions, specified below.
{1) \textbf{Zero-shot learning:} previous works have leveraged pre-trained VLMs for zero-shot depth estimation \cite{zhang_can_2022}, object counting \cite{jiang_clip-count_2023}, \etc For example, \cite{zhang_can_2022} utilizes the pre-trained knowledge from CLIP to project the semantic response of each image patch into a certain depth bin, and linearly combine depth values in the bin to obtain the final depth predication. Similarly, we can define the frequency bin (\eg, [0.5, 1.0, 1.5, 2.0, 2.5]) and use every value in the bin to construct text prompt describing the frequency of the skin color temporal variation, such as "the frequency of the skin color variation is 2.0 hertz in the video." Next, we calculate the similarity between the visual embedding of the given video and the textual embedding of the text prompt via the VLM. Finally, 
% the text prompt with the highest similarity score is regarded as the main frequency for rPPG signal generation. 
we compute the weighted sum of the values in the frequency bin and regard it as the main frequency for the rPPG signal.
However, pre-trained VLMs excel in extracting temporal features but struggle to understand the frequency-related  attributes. This is due to the fact that the pre-trained datasets for VLMs rarely contain vision-text pairs that describe periodic events. We thereby decide not to expand along this line.

2) \textbf{Fine-tuning:} some studies focus on fine-tuning VLMs for specific downstream tasks. Due to the lack of text annotations in the datasets, they typically begin with some vision-text pair generation methods and subsequently  optimize the models through multimodal representation learning. For example, 
% \cite{liang_crowdclip_2023} constructed ranking text prompts to describe the crowd counts within different patches in the given image. Then they facilitate the vision encoder in capturing crowd semantics for crowd counting. 
% Chatterjee \etal 
\cite{liang_crowdclip_2023} manually constructs text prompts to describe the crowd counts within patches of different sizes in the given image. Then a multimodal ranking loss is devised to facilitate the vision encoder of the VLM to capture the crowd semantics in the image for crowd counting.
% \cite{chatterjee_robustness_2024} generates text prompts to describe the spatial relationships (in front of, above, far away, etc.) of objects within images
% to help the vision encoder understand the spatial relationships (in front of, above, far away, etc.) of objects within images for monocular depth estimation.
This research line is feasible if we have the actual frequency of the rPPG signal to describe the facial video, which is however undesirable in our self-supervised remote physiological measurement task. 
% Instead, we pose a new question: is it possible to generate the text prompts while bypassing the ground truth signal frequency?

% , and instead based on the comparison (\ie, relative ratio) of the signal frequencies between different videos?
% create text descriptions corresponding to the comparison (\ie, relative ratio) of the signal frequencies between different videos?

% In this paper, we propose a pioneering approach, VL-phys, to bootstrap VLMs for self-supervised rPPG estimation. As shown in Fig.~\ref{fig1}, we introduce a novel vision-text pair generation method where facial videos are augmented with well-designed rPPG frequency relations while text prompts are generated to describe such relations. Afterwards, we fine-tune the pre-trained VLM with these vision-text pairs to enhance its ability to capture frequency-related attributes for rPPG estimation.

% which creates the text descriptions for the vision modality based on the comparison (\ie, relative ratio) of the signal frequencies between different videos. 原来
% which creates the text descriptions for the vision modality based on the comparison (\ie, relative frequency ratio) of the facial color changes between different videos. 是不是也可以呢？

In this paper, we introduce a pioneering approach to bootstrap VLMs for self-supervised remote physiological measurement. Our proposed method, namely VL-phys, marks the first attempt to adapt VLMs with the ability to digest the frequency-related knowledge in vision and text modalities. As shown in Fig.~\ref{fig1}, we introduce a novel frequency-oriented generation mechanism to create vision-text pairs reflecting the frequencies of skin color temporal variations. Given the input facial video, we first augment it into multiple positive and negative samples with their rPPG frequencies conforming certain relations; next, in the absence of their ground truth signal frequencies, we create contrastive spatio-temporal maps between positive and negative samples and craft text prompts to describe the frequency relations across samples; this forms vision-text pairs. Afterwards, we fine-tune the pre-trained VLM with these created vision-text pairs through well-designed {frequency-related} representation learning in both generative and contrastive manners. The former features a text-guided visual reconstruction task in which we reconstruct masked patches of contrastive spatio-temporal maps under the guidance of text prompts; while the latter consists of a series of tasks, such as the multimodal vision-text contrastive learning task to align the embeddings of vision-text pairs, and the unimodal frequency contrastive and ranking task to optimize rPPG signals estimated from visual embeddings of different video samples.

Overall, the contribution of this work can be summarized in four-fold:

\begin{itemize}
\item  We propose VL-phys, a novel {frequency-centric} self-supervised framework for remote physiological measurement based on the pre-trained VLM. It innovatively adapt the pre-trained VLM to 
%\st{encode periodic pulsation information in facial videos via multimodal contrastive learning} 
digest the frequency-related knowledge in vision and text modalities via 
% multimodal 
representation learning in both contrastive and generative manners.
% in periodic color variations.

\item We introduce a frequency-oriented vision-text pair generation mechanism to enable the
% multimodal 
representation learning.
%Instead of creating text descriptions for the vision modality based on the 
Without relying on the ground truth PPG signals, we carefully augment facial videos and create contrastive spatio-temporal maps to reflect the frequency ratios of skin color temporal variations across different samples; an appropriate text prompt is then designed to 
describe such relative frequency relation in these visual maps.
% the relative frequency ratios  
%These texts can guide the vision encoder to learn the signal frequency differences between videos.}

\item We design a multitask learning pipeline that incorporates the multimodal generative and contrastive tasks (\ie, the text-guided visual reconstruction task and the vision-text contrastive learning task) to help optimize the self-supervised rPPG estimation. Besides the multimodal learning, we also introduce the unimodal frequency contrastive and ranking task to constrain rPPG estimation among multiple augmented video samples.

% to enhance the vision encoder to capture frequency-related visual attributes for rPPG estimation.

\item We conduct extensive experiments on four publicily avaiable datasets, UBFC-rPPG \cite{bobbia_unsupervised_2019}, PURE \cite{stricker_non-contact_2014}, VIPL-HR \cite{niu_rhythmnet_2020} and MMVS \cite{yue_deep_2021}; our results demonstrate that VL-phys significantly outperforms state of the art self-supervised methods, and performs even on par with state of the art supervised methods.
\end{itemize}

Our method, beyond its merits in the remote physiological measurement task, should also benefit many other tasks that desire VLMs to understand the periodic and frequency-related attributes in their modalities, such as gait recognition \cite{liu2022monitoring}, weather change prediction \cite{sonderby2020metnet}, Parkinson’s disease diagnosis \cite{yang2022artificial}.

\medskip 

%To the best of our knowledge, it is the first time to leverage \st{the linguistic cues from} text prompts to enhance the vision encoder's understanding of frequency-related visual attributes.

\section{Related work}

\subsection{Remote physiological measurement}
Traditional facial video-based remote physiological measurement methods normally estimate rPPG using the blind signal separation techniques \cite{wang_algorithmic_2017,lam_robust_2015} or skin reflection models \cite{de_haan_robust_2013,wang_novel_2016}. First, based on assumption that the temporal variation of skin color is a linear combination of the target rPPG signal and other irregular noises, blind signal separation-based approaches leverage independent component analysis (ICA) or principal components analysis (PCA) to disentangle rPPG signals from noises. For example, 
% McDuff \etal \cite{mcduff_fusing_2018} utilized ICA to decompose signals and estimate rPPG from facial videos captured at various angles. Similarly, 
Lam \etal \cite{lam_robust_2015} applied ICA to several non-overlapping video patches for their independent rPPG estimation. They proposed a frequency-based majority voting strategy to determine the most accurate estimation. Second, skin reflection model-based approaches aim to project the original images into different color spaces that can better highlight rPPG components. For instance, Haan \etal \cite{de_haan_robust_2013} proposed 
% a color space projection approach 
CHROM to project from the RGB space to the chrominance space for reducing the motion noises and improving rPPG estimation. 
% Wang \etal \cite{wang_novel_2016} proposed a spatial subspace rotation (2SR) approach to further enhance the robustness against motion and illumination variations. 
However, these two kinds of traditional approaches are often designed under prior assumptions, \eg, they assume that different individuals have uniform skin tones under white light \cite{de_haan_robust_2013}, which may not satisfy in complex situations, hence potentially impeding the rPPG estimation.

Recently, deep learning approaches show remarkable performance in rPPG estimation \cite{yu_physformer_2022,lu_dual-gan_2021,qiu_evm-cnn_2019,vedaldi_meta-rppg_2020,yu_remote_2019,niu_video-based_2020,li_learning_2023,wang2022self,li2023contactless}. For example, Niu \etal \cite{niu_video-based_2020} proposed a feature disentangling mechanism that effectively separates physiological features from noises by employing an encoder-decoder network. Yu \etal \cite{yu_physformer_2022} utilized self-attention layers to capture long-range spatio-temporal relationships among multiple video clips for rPPG estimation. Gupta \etal \cite{gupta_radiant_2023} based their work on vision transformer to regress rPPG waveforms. 
% Du \etal \cite{du_dual-bridging_2023} improved the generalizability of rPPG estimation methods to unseen scenarios by using a generative adversarial network. Moreover, Yu \etal \cite{yu_autohr_2020}, Lee \etal \cite{lee_meta-rppg_2020} and Yin \etal \cite{yin_pulsenet_2022} further improved the estimation accuracy and robustness through neural architecture search, meta-learning, and multi-task learning mechanisms, respectively. 
Despite their remarkable achievements, these supervised learning approaches demand extensive facial videos and synchronously recorded PPG signals for model training. The collection of PPG signals is both costly and time-consuming. Consequently, there has been a recent surge in interest to optimize the model using contrastive self-supervised learning \cite{yang_simper_2023,yue_facial_2023,gideon_way_2021,sun_contrast-phys_2024}. Specifically, they first apply spatial/frequency augmentations to create the positive/negative samples for the given video. Next, they pull close the predicted rPPG signals from positive samples, and repel the predictions from positive samples to negative samples.
% Among them, a popular solution is to first apply spatial augmentations on the given video to create its positive counterparts, meanwhile its resampled/frequency-augmented versions, or other videos in the training batch, are considered as the negative samples, and finally a contrastive loss is applied among rPPG signals from the positive and negative samples for network optimization \cite{yang_simper_2023,yue_facial_2023,gideon_way_2021,sun_contrast-phys_2022}. 
For instance, Yue \etal \cite{yue_facial_2023} designed a frequency augmentation module to modify the frequency of the rPPG signal for the given video and applied contrastive learning between the transformed and original videos.
Similarly, Sun \etal \cite{sun_contrast-phys_2024} presented Contrast-Phys+ which contrasts rPPG similarity among different spatial crops and temporal clips within the given video. Due to the absence of supervision from PPG signals, learning the periodic temporal information is still very challenging for the self-supervised learning. 
%leading their performance inferior to state of the art supervised-learning methods.

% To address this limitation, unlike existing works that solely contrast visual embeddings encoded from videos, we extend our previous approach \cite{yue_facial_2023} by incorporating text supervision to facilitate multimodal contrastive learning within the self-supervised framework for facial video-based remote physiological measurement. Specifically, we construct text prompts to describe the relative magnitudes of signal frequencies across different video samples, and generate the vision-text pairs for multimodal feature alignment. We leverage the linguistic cues from text descriptions to enhance the vision encoder to extract more discriminative temporal features for accurate rPPG estimation.

% incorporate text supervision to facilitate self-supervised rPPG estimation.
% to enhance the vision encoder to extract more discriminative pulsation information for accurate estimation
% \ZJ{To tackle above, we \st{propose a vision-text pair generation method and} exploit the \st{linguistic cues from} text descriptions to enhance the vision encoder for more accurate rPPG estimation.}
To tackle above, we exploit text prompts to help capture frequency-related visual attributes in videos for accurate rPPG estimation. We are the first to introduce VLM into the remote physiological measurement task. %It is worth noting that leveraging the multimodal contrastive learning and VLMs for remote physiological measurement is unique in both self-supervised and supervised rPPG estimation.

\begin{figure*}[!t]
  \centerline{\includegraphics[width=6in]{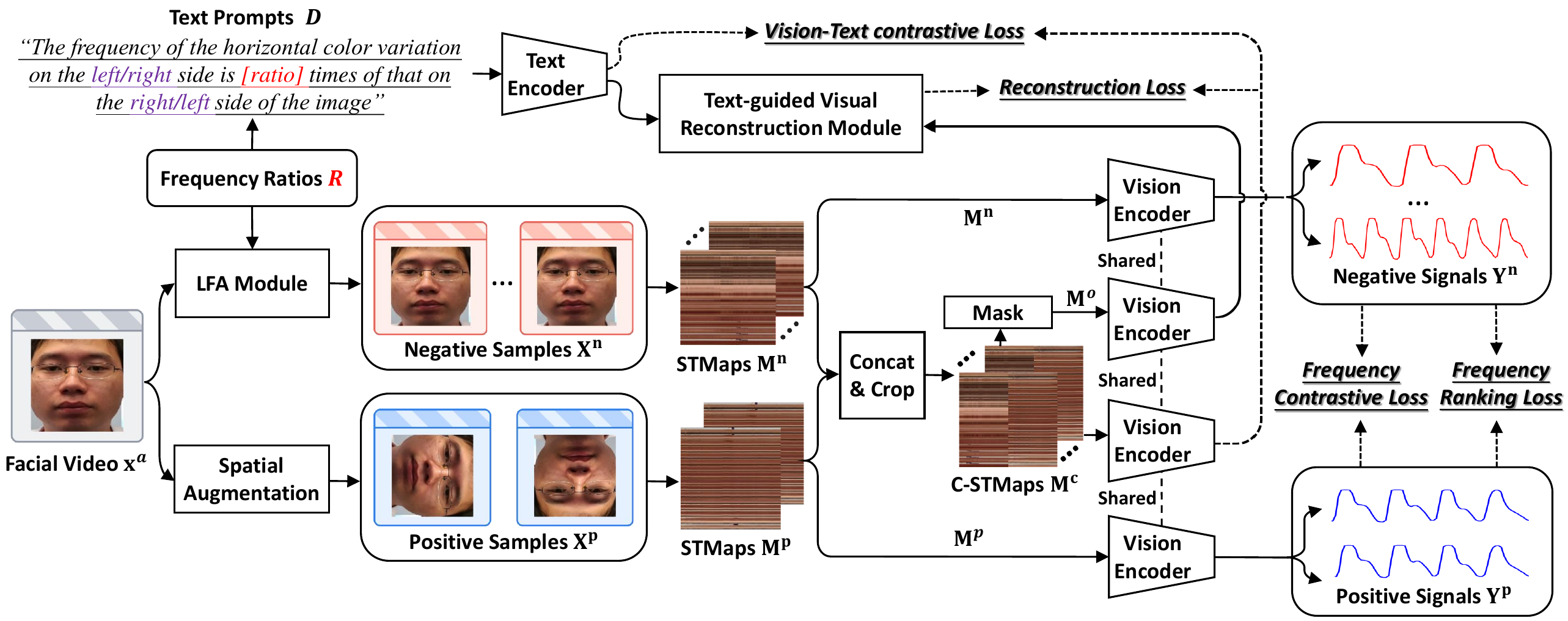}}
  %\vspace{-0.1in}
  \caption{Overall architecture of our VL-phys. Given an input video, we respectively apply spatial augmentation and learnable frequency augmentation (LFA) to obtain its positive and negative video samples. We generate their spatio-temporal maps (STMaps) and then create contrastive spatio-temporal maps (C-STMaps) to reflect the frequency ratios of skin color temporal variations between positive and negative samples; meanwhile, we carefully craft text prompts to describe such relations. Afterwards, we fine-tune the pre-trained vision and text encoders of VLM with these formed vision-text pairs via {frequency-related} multimodal generative and contrastive tasks, \ie the text-guided visual reconstruction task and vision-text contrastive learning task. Moreover, we introduce the unimodal frequency contrastive loss and the frequency ranking loss to optimize the rPPG signals estimated from different video samples.}
  % further improve the vision encoder ability to capture frequency-related visual attributes for rPPG estimation.}
  \label{fig2}
  % \vspace{-0.1in}
  \end{figure*}

% It comprises five main stages: 1) data augmentation. 2) frequency-oriented vision-text pair generation, 3) feature extraction, 4) visual feature reconstruction, 5) network optimization.

\subsection{Vision-Language Model}
Vision-language models (VLMs) such as CLIP \cite{radford_learning_2021}, BLIP \cite{li_blip_2022}, VideoCLIP \cite{xu_videoclip_2021}, and MiniGPT4 \cite{zhu_minigpt-4_nodate} have garnered significant attentions in recent years due to their impressive performance in both multimodal and unimodal downstream tasks. They learn cross-modal representations by aligning vision and text modalities through extensive training on a vast corpus of vision-text pairs. Building upon CLIP, Ye \etal \cite{ye_hitea_2023} proposed a cross-modal moment exploration task and a multi-modal temporal relation exploration task to enhance the temporal feature modelling ability of their encoders.
% to model the temporal contextual relations among different views of the video-text pair 
It achieves state of the art results on downstream tasks like videoQA, video captioning and video-text retrieval. Varma \etal \cite{varma_villa_2023} decomposed the image-text samples into multiple region-attribute pairs, and proposed ViLLA to capture fine-grained relationships between image regions and textual attributes. Moreover, for specific visual tasks where off-the-shelf text descriptions are unavailable, researchers typically synthesize text prompts to enable vision-text contrastive learning. For instance, Chatterjee \etal \cite{chatterjee_robustness_2024} created text prompts 
to help the vision encoder understand the spatial relationships (in front of, above, far away, \etc) among 
objects in images for monocular depth estimation. Similarly, in the domain of surgical instrument segmentation, Zhou \etal \cite{zhou_text_2024} generated text prompts for instruments using outputs by the large language model GPT-4. In this paper, we novelly construct text prompts to describe the frequency ratios of skin color temporal variations over different video samples, and eventually align them with the vision modality in the embedding space to solve the rPPG estimation task.

\section{Method}

\subsection{Overview}\label{sec3.1}

The method overview is illustrated in Fig.~\ref{fig2}. Our goal is to bootstrap and finetune VLM to learn rPPG signals from unlabelled facial videos using 
% both unimodal and
{frequency-related} generative and contrastive learning mechanisms. The framework contains five phases: data augmentation, vision-text pair generation, feature extraction, visual reconstruction and network optimization.

Specifically, given a facial video $x^a$, we first 
% use the open-source face detector 3DDFAv2 \cite{guo2020towards} to detect, align and crop the facial ROIs in each frame. Then in the data augmentation stage, we
perform spatial/frequency augmentation on it to create a number of positive/negative samples. The positive samples $X^p=\left \{ x^p \right \}$ maintain the same rPPG signal frequency ($f^a$) to $x^a$, while the rPPG signal frequencies of negative samples are transformed to $r \times f^a$ by feeding a set of frequency ratios $R=\left \{ r \right \}$ into a learnable frequency augmentation (LFA) module. Then we generate the spatio-temporal maps (STMaps) $M^p=\left \{ m^p \right \}$ and $M^n=\left \{ m^n \right \}$ for both positive and negative samples. These STMaps represent the temporal variations of skin colors in facial ROIs. Next, we introduce a novel frequency-oriented vision-text pair generation method: we create a set of contrastive spatio-temporal maps (C-STMaps) $M^c=\left \{ m^c \right \}$ by concatenating any $m^p$ with $m^n$ horizontally. Since the frequency ratio of rPPG signals between $m^p$ with $m^n$ is a fixed value $r$, we construct the text prompt $d\in D$ that depicts the relative frequency ratio of the color variation from left to right in $m^c$. By this means, $m^c$ and $d$ form a vision-text pair. Meanwhile, several non-overlapping patches of $m^c$ are randomly masked to generate the masked contrastive spatio-temporal maps (M-STMap) $m^o \in M^o$. % $M^o=\left \{ m^o \right \}$. 
Having $M^p$, $M^n$, $M^c$ and $M^o$, in the third stage, we pass them into the vision encoder to 
% \st{encode their pulsation information and} 
obtain visual embeddings $V^p$, $V^n$, $V^c$ and $V^o$. Meanwhile, the text prompts in $D$ are input into the text encoder to obtain textual embeddings. We leverage the pre-trained vision and text encoders from the popular VLM, VindLU \cite{cheng_vindlu_2023}. After that, an rPPG estimation head is applied to project $V^p$, $V^n$
% =\left \{v^p \right \}$, $V^n=\left \{v^n \right \}$ 
into rPPG signals $Y^p=\left \{y^p \right \}$ and $Y^n=\left \{y^n \right \}$. In the visual reconstruction stage, we propose a text-guided visual reconstruction (TVR) module to reconstruct the masked patches in $m^o$ under the guidance of text prompts.
%which implicitly enforces the vision encoder to gain a deeper understanding of the periodic color variation in XX. 
Last, in the main network optimization stage, the vision-text contrastive loss \cite{radford_learning_2021} is applied to align the embeddings of every vision-text pair $m^c$ and $d$. The frequency contrastive loss \cite{yue_facial_2023} is exploited among the power spectral densities (PSD) of signals $Y^p$ and $Y^n$. Also, we introduce a novel frequency ranking loss among $Y^p$ and $Y^n$ to let the vision encoder accurately predict their frequency ranks.

\subsection{Data augmentation}\label{sec3.2}

Given the input facial video $x^a$, we apply spatial/frequency augmentation strategies on it to generate positive/negative samples, respectively. 

The spatial augmentations (\ie rotations, horizontal and vertical flips) change the spatial layouts of the video frames without altering their colors. Consequently, the frequency of the rPPG signal inside $x^a$ remains unaffected. In practice, we randomly choose two augmentation operations to create two positive samples $X^p = \{x^p_1, x^p_2\}$ for $x^a$. The main frequency of their rPPG signals is the same to that ($f^a$) of $x^a$.

To obtain negative samples, we apply the learnable frequency augmentation (LFA) module \cite{yue_facial_2023} on $x^a$ to manipulate its rPPG signal frequency with target ratios. Specifically, we randomly sample three frequency ratios $R = \{r_1, r_2, r_3\} $ within a ratio bin [0.25,0.5,0.75,1.25,1.5,1.75,2] and feed them into the LFA module along with $x^a$ to create three negative samples $X^n = \{x^n_1, x^n_2, x^n_3\}$. 
%\ZJ{(experiments on different number of negative samples are presented in Sec.~\ref{sec6.7}).}
Although $X^n$ retain the general visual appearance of $x^a$, the main frequencies of their rPPG signals are modulated to $\{r_1 \times f^a, r_2 \times f^a, r_3 \times f^a\}$. %This is achieved via the frequency modulation block of the LFA module, which performs non-linear signal frequency transformation on multi-scale features of $x^a$. 
% To elaborate, the frequency modulation block initially estimates a rough rPPG signal of $x^a$ and predicts a frequency modulation vector using 1D Res-blocks and Bidirectional LSTM layer. This modulation vector is then element-wisely multiplied with the feature of $ x^a$ to change its rPPG signal frequency from $f^a$ to $r \times f^a $. Finally, the LFA module reconstructs the video clip $x^n$, which is regarded as a negative sample to $x^a$. 

Next, we leverage a face detector (3DDFAv2 \cite{guo2020towards}) to detect pulsation-sensitive facial ROIs (\eg, forehead, nose) while masking out other areas (\eg, backgrounds) in these samples. Following \cite{niu_video-based_2020,niu_rhythmnet_2020,lu_dual-gan_2021}, we transform positive samples $X^p$ and negative samples $X^n$ into spatio-temporal maps (STMaps) $M^p = \{m^p \in \mathbb{R}^{A \times F \times C}\}$ and $M^n = \{m^n \in \mathbb{R}^{A \times F \times C}\}$. Each row of STMap describes the temporal variation of the skin color within a certain facial ROI. $A$ denotes the number of facial ROIs, $F$ denotes the frame number of the video clip, and $C$ denotes the channel number of color space ($C=3$ for the RGB space). STMaps collapse the original 4-D video features into the 3-D format, which not only highlights the skin color temporal variations in facial videos but also accelerates model training and inference. Afterwards, we resize each $m$ into $\mathbb{R}^{F \times F \times C}$ to ensure the compliance with the input size requirement for the VLM.
% This way collapses the original 4-D video into 3-D STMaps, which is also good for accelerating the model training and inference.

%The STMap emphasizes the skin color temporal variations from facial ROIs. 
% , and represents the color variation signals in a single image
% Each row of STMap mi represents the raw temporal signal
% for one ROI on the face.

\subsection{Vision-text pair generation}\label{sec3.3}

\begin{figure}[!t]
  \centerline{\includegraphics[width=3.6in]{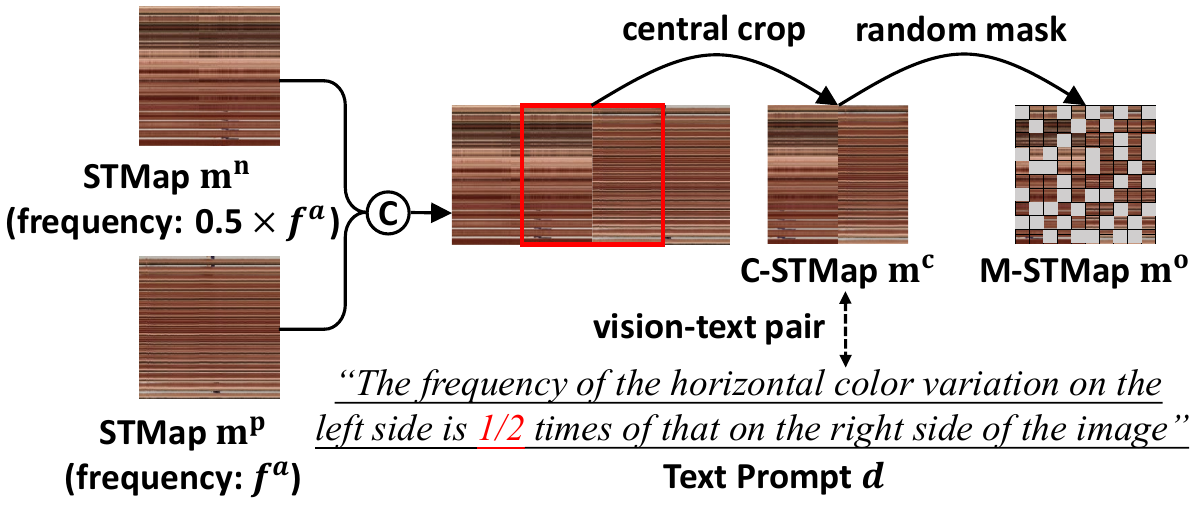}}
  %\vspace{-0.1in}
  \caption{The process of the generation of frequency-oriented vision-text pair and masked contrastive spatio-temporal map (M-STMap).}
  \label{fig3}
  % \vspace{-0.1in}
  \end{figure}

We aim to fine-tune the VLM to capture frequency attributes of $M^p$ and $M^n$ for rPPG estimation via representation learning. However, neither the ground truth signals or off-the-shelf text prompts are available in the self-supervised setting. Fortunately, the signal frequency between negative and positive samples are fixed ratios, \ie, the frequency of the horizontal color variation in the negative STMap $m^n$ is $r$ times of that in the positive STMap $m^p$. We can design text prompts to illustrate their frequency relations. 
% \emph{XXXXXX}. 
Below we specify the details. 

\subsubsection{Contrastive spatio-temporal map generation}\label{sec3.3.1}

% Given STMaps $M^p = \{ m^p \in \mathbb{R}^{F \times F \times C} \} $ for the positive samples and $M^n = \{ m^n \in \mathbb{R}^{F \times F \times C} \} $ for the negative samples, \ZJ{we perform horizontal concatenation on each pair of $m^p$ and $m^n$ to create a set of contrastive spatio-temporal maps (C-STMaps) $M^c = \{ m^c_j \in \mathbb{R}^{F \times F \times C} | j = 1, 2, \ldots, 6 \}$. 
Given two positive STMaps $M^p = \{ m^p_1, m^p_2 \} $ and three negative STMaps $M^n = \{ m^p_1, m^n_2, m^n_3 \} $, we perform horizontal concatenation between every $m^p$ and every $m^n$ to create a set of contrastive spatio-temporal maps (C-STMaps) $M^c = \{ m^c_j \in \mathbb{R}^{F \times F \times C}| j \in \left [1,6 \right ]  \}$. For example, as depicted in Fig.~\ref{fig3}, supposing $m^n$ contains an rPPG signal with a frequency of $0.5 \times f^a $ and $ m^p $ maintains an rPPG signal with a frequency of $f^a$. We horizontally concatenate them 
% to obtain $m^r \in \mathbb{R}^{F \times 2F \times C}$. Subsequently, we 
%In this configuration, the left side of $ m^r $ corresponds to $m^n $ with the horizontal color variation of the frequency $0.5 \times f^a $, while the right half side is derived from $m^p $ with the horizontal color variation of the frequency $f^a $. 
and execute a central crop operation to produce C-STMap $m^c$. The left and right side of $m^c$ respectively reflects the temporal variation of skin color in the positive and negative STMap, $m^p$ and $m^n$. Therefore, the frequency of the horizontal color variation in its left side is $r=\frac{1}{2}$ times of that in its right side.

Note that here we use the horizontal concatenation to ensure that each row corresponding to the same ROI in the positive and negative STMaps are properly aligned. This process yields the C-STMap in which the frequency of horizontal color variation displays a noticeable change at the boundary of the two sides. This arrangement allows us to easily design text prompts to  describe the relative ratio of signal frequencies between the left and right sides of C-STMap. In contrast, using other concatenation methods, such as vertical or channel concatenation, would alter the spatial alignment between the positive and negative STMaps, and the signal frequency difference would not be manifesting to the same extent as along the single horizontal axis of C-STMap. This would make it even undesirable to generate corresponding text prompts for the frequency-related representation learning.

% pulsation information from $m^p$ and $m^n$.

%\ZJ{Moreover, such as arrangement enables the subsequent vision encoder to directly capture their frequency differences from $m^c$.}
% Moreover, since $m^p$ and $m^n$ exhibit different color variation frequencies, this arrangement allows the subsequent vision encoder to analyze and distinguish their frequency differences from $m^c$ via the feature interactions facilitated by self-attention layers.

\subsubsection{Frequency-oriented vision-text pair generation}\label{sec3.3.2}
Next, we generate text prompt $d \in D$ to respectively describe the relative ratio of signal frequencies between the left and right sides of each $m^c$. Specifically, $d$ is defined as: 
% “the frequency of skin color temporal variation on the left/right side is [ratio] times of that on the right/left side of the image”.
“the frequency of the horizontal color variation on the left/right side is [ratio] times of that on the right/left side of the image”.
Here, the [ratio] token is replaced with the specific frequency ratio $r$ between negative and positive samples. For the example shown in Fig.~\ref{fig3}, the [ratio] token is replaced with the value “1/2”.
% By this means, the C-STMaps $M^c = \{ m^c_j \}$ and text descriptions $D = \{ d_j \} $ effectively form the vision-text pairs for multimodal contrastive learning. 
By this means, the C-STMap $m^c_j$ and its corresponding text prompt $d_j$ effectively form a positive vision-text pair. In contrast, $m^c_j$ combined with other C-STMaps' text prompts $\{d_l| l \in \left [1,6  \right ] , l \ne j\}$ form negative vision-text pairs. These pairs are used for the subsequent vision-text contrastive learning and text-guided visual reconstruction. The experimental analysis of other text templates is presented in Sec.~\ref{sec6.4}.

% Notably, we convert the Arabic numeral $r $ into its English word form before inserting it as the [rat] token. This conversion is based on our observation that the text encoder of BLIP understands the semantics of words better (Sec.~\ref{sec5.10}).

\subsection{Feature extraction}\label{sec3.4}

In this stage, we extract features from the generated vision-text pairs. We encode C-STMaps $M^c = \{ m^c_j \}$ and their associated text prompts $D = \{ d_j \}$ into visual embeddings $V^c = \{ v^c_j \}$ and textual embeddings $E = \{ e_j \}$ using the pre-trained vision and text encoders from VindLU \cite{cheng_vindlu_2023}, respectively.
% perform multimodal contrastive learning based on the
% Importantly, we utilize the robust capabilities of the vision and text encoders from the pre-trained VindLU \cite{cheng_vindlu_2023}. Then, we further improve the vision encoder's understanding of frequency patterns in periodic color variations, guided by the textual descriptions provided by $D$.
% Then, we align them using the vision-text contrastive loss

\textbf{Vision encoder:} The vision encoder is a vision transformer that processes the input $m^c$ by dividing it into patches and encoding them as a sequence of embeddings. It includes an additional [CLS] token that represents the global feature of $m^c$. Positional embeddings are added to the patch embeddings to retain positional information, which is crucial for the encoder to understand the spatial layout of $m^c$. We use $v^c$ to denote the obtained visual embedding, and specifically, $\hat{v}^c$ the embedding of the [CLS] token.
% \in \mathbb{R}^{1 \times B}$, where $B$ is the feature dimension.

\textbf{Text encoder:} The text encoder has the same structure as BERT \cite{devlin_bert_2019}. We feed the tokenized text prompt $d$ into it for feature extraction. We append a [CLS] token to the beginning of $d$ as a global summary of the prompt. The embedding of this [CLS] token is denoted as $\hat{d}$ 
% \in \mathbb{R}^{1 \times B}$ 
and is treated as the global textual embedding of $d$.

After obtaining  $\hat{v}^c$ and $\hat{d}$ for each vision-text pair, we fine-tune the encoders by encouraging the feature similarity between $\hat{v}^c$ and $\hat{d}$ via the vision-text contrastive loss (Sec.~\ref{sec3.6}, $L_{vtc}$).

Moreover, we employ the vision encoder to extract the embeddings of the global [CLS] tokens from two positive STMaps $M^p=\{m^p_1, m^p_2\}$ and three negative STMaps $M^n=\{m^n_1, m^n_2, m^n_3\}$, \ie, $\hat{v}^p_1$, $\hat{v}^p_2$, $\hat{v}^n_1$, $\hat{v}^n_2$, $\hat{v}^n_3$. Then they are respectively fed into rPPG estimation head (\ie a linear layer) to project them into positive signals $Y^p = \{ y^p_1,y^p_2 \}$ and negative signals $Y^n = \{ y^n_1,y^n_2,y^n_3 \}$.
% visual embeddings $V^p$ and $V^n$ 
%\ZJ{(The experimental analysis between the global [CLS] token and local image patches is presented in Sec.~\ref{sec6.6}).}
These signals are of various frequencies and can be used for unimodal frequency contrastive learning (Sec.~\ref{sec3.6}, $L_{fc}$).
% self-supervised training

Subsequently, we concatenate these predicted signals $Y = \{ y^p_1,y^p_2,y^n_1,y^n_2,y^n_3 \}$
% along the temporal dimension 
and apply a linear layer to predict their frequency ranks. We optimize the predicted ranks against the ground truth ranks (Sec.~\ref{sec3.6}, $L_{fr}$).

\begin{figure}[!t]
  \centerline{\includegraphics[width=1.6in]{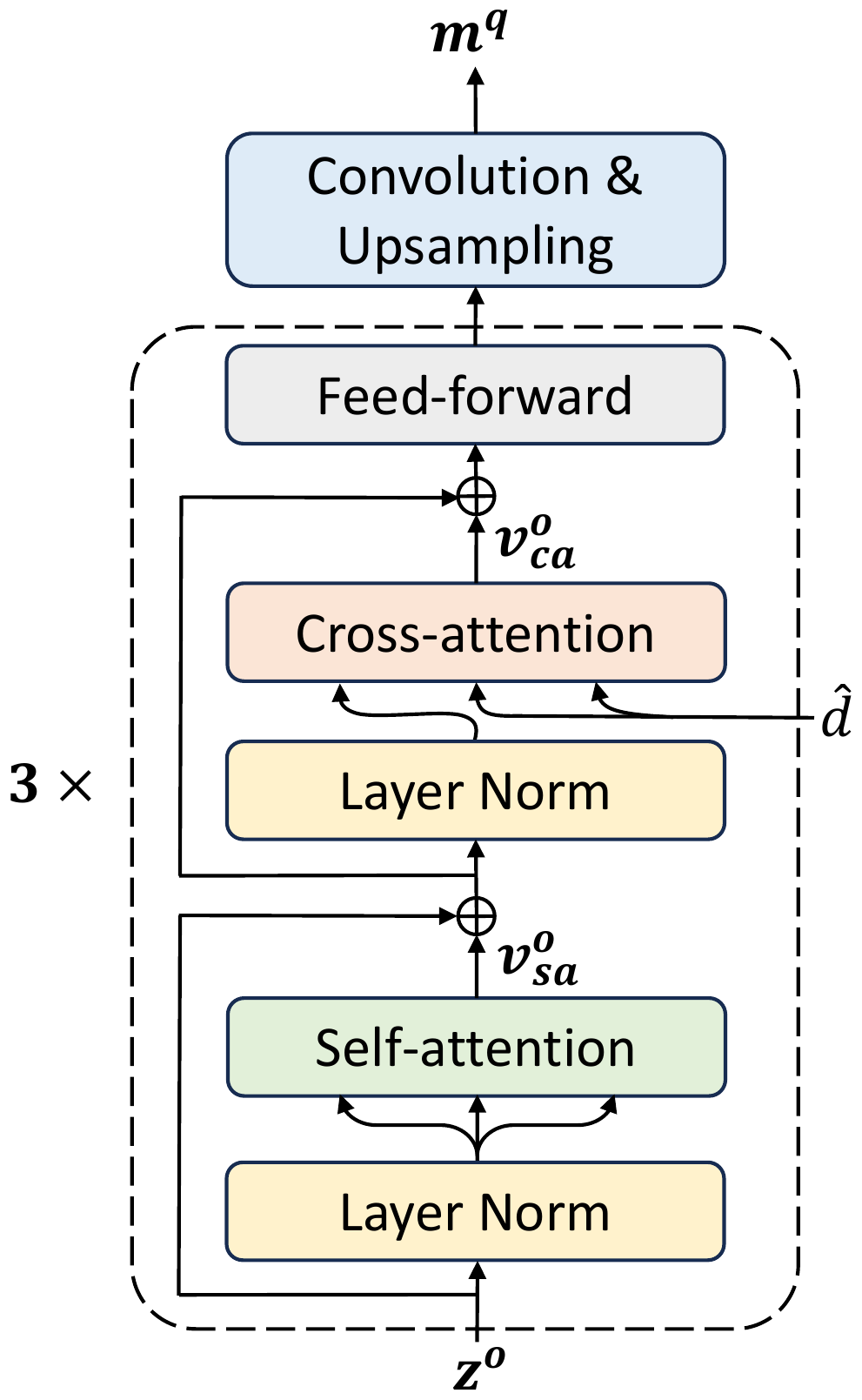}}
  %\vspace{-0.1in}
  \caption{The structure of text-guided visual reconstruction (TVR) module.}
  \label{fig4}
  % \vspace{-0.1in}
  \end{figure}

\subsection{Visual reconstruction}\label{sec3.5}

%It has been demonstrated in VLMs that text-guided masked image modeling task can effectively encourage the encoders to align the embeddings between vision and text modalities \cite{bao_vl-beit_2022,kwon_masked_2023}. 
% For instance, if the text describes the objects that are not visible due to masking, the encoder learns to infer these objects' presence and characteristics, which enhances its predictive capabilities. 
In this stage, we enable the encoders to develop a deeper understanding of the frequency-related attributes via a {frequency-related} generative learning mechanism.
%\st{This stage aims at recovering masked image patches based on linguistic cues from texts, enabling the vision encoder to develop a deeper contextual understanding of images. To implement it, we leverage an auxiliary task to} 
%\ZJ{We implement this auxiliary task} 
%\st{enforce our vision encoder to capture discriminative periodic or frequency features of visible patches, enabling the reconstruction of} 
We develop a text-guided visual reconstruction task to
reconstruct masked patches of C-STMaps with the guidance of the text prompts.
% on the skin color temporal variations %in the images. 
This task can implicitly enhance our vision encoder to learn the periodicity of rPPG signals. Below we specify the details.

\subsubsection{Masked contrastive spatio-temporal map generation}\label{sec3.5.1}
As shown in Fig.~\ref{fig3}, we first divide each C-STMap $m^c$ into non-overlapping patches and randomly mask the patches with a ratio of $b$ to the total number of patches. The visible patches in the masked contrastive spatio-temporal map (M-STMap) $m^o$ are then fed into the vision encoder to obtain the visual embedding $v^o$.
% $V^o=\left \{v^o \right \}$.

\subsubsection{Text-guided visual reconstruction}\label{sec3.5.2}

We further design a text-guided visual reconstruction (TVR) module to recover the masked patches in $m^o$. Specifically, the TVR module contains three interaction blocks. Its structure is illustrated in Fig.~\ref{fig4}. The input for the TVR module, $z^o$, is the concatenated sequence of encoded visible patch features $v^o$ and the learnable masked tokens (with positional embeddings). Each interaction block begins by calculating the self-attention (SA) within $z^o$. It is expressed as: $v_{sa}^o= SA(LN(z^o))+z^o$, where $LN(.)$ represents the layer normalization. Then we apply the cross-attention (CA) between $v_{sa}^o$ and its associated 
% global 
textual embedding $\hat{d}$ to facilitate their interaction and obtain joint representation $v_{ca}^o$; $v_{sa}^o$ is treated as the query and $\hat{d}$ is treated as both the key and value for CA. This process is written as: $v_{ca}^o  =CA(LN(v_{sa}^o),\hat{d})+v_{sa}^o$. Subsequently, a feed-forward layer is exploited to further refine $v_{ca}^o$. After executing three such blocks, we reconstruct the patches and map the output back to the original RGB space via a $1\times1$ convolution layer and an upsampling layer.

\subsubsection{Reconstruction loss}\label{sec3.5.3}

We leverage the reconstruction loss $L_r$ to minimize the distance between the reconstructed and original C-STMaps. It is formulated as:
\begin{equation}\label{equ1}
L_r= \frac{1}{6}\sum_{j=1}^{6}MSE(m^c_j,m^q_j)
\end{equation}
where $MSE(.)$ denotes the mean squared error, $m^q$ represents the reconstructed visual map.

% This loss activates the TMR module, facilitating the fine-grained alignment between vision and text according to the semantics related to frequency. Moreover, it implicitly ensures that the vision encoder develops a deeper understanding of the dynamics within the visual modality, and compels the vision encoder to accurately capture the periodic patterns of the color variations for improving rPPG estimation.

The text prompt $d$ specifies the frequency relation of the color variation between the left and right parts of $m^o$. It serves as guidance for the vision encoder to capture the periodic color variation patterns of visible patches in $m^o$, facilitating the correct reconstruction of the remaining masked patches.

%. Based on it and the embeddings of visible tokens in $m^o$, the TMR module reconstructs the complete image corresponding to the text. \ZJ{This task implicitly optimizes the vision encoder to develop a deeper understanding of frequency-related visual attributes, and enforces it to capture periodic patterns of horizontal color variations, otherwise we cannot reconstruct correct images.}

%\ZJ{The experimental analysis of another similar auxiliary task, \ie, image-guided masked language modeling task, is presented in Sec.~\ref{sec6.2}.}

\subsection{Network optimization}\label{sec3.6}

Besides the reconstruction loss $ L_{r} $, we introduce the vision-text contrastive loss, the frequency contrastive loss, and the frequency ranking loss for network optimization.

\textbf{Vision-text contrastive loss} ($L_{vtc}$): This loss aims to align text with vision in the embedding space. This is achieved by encouraging the similarities between visual and textual embeddings from positive vision-text pairs (\ie $m^c_j$ and $d_j$) while distinguishing them from negative pairs (\ie $m^c_j$ and \{$d_l\}$) (see Sec.~\ref{sec3.3.2}). We define it as:

% \begin{align}\label{equ2}
% % \scriptsize
% L_{vtc} = -\frac{1}{J}\sum_{j=1}^{J}\log [\frac{\exp(\left \langle\hat{v}^c_j, \hat{d}_j\right \rangle / \tau_1)}{\sum_{k=1,k\ne j}^{J}\exp(\left \langle\hat{v}^c_j, \hat{d}_k\right \rangle / \tau_1)} \\
% +\log \frac{\exp(\left \langle\hat{v}^c_j, \hat{d}_j\right \rangle / \tau_1)}{\sum_{k=1,k\ne j}^{J}\exp(\left \langle\hat{v}^c_k, \hat{d}_j\right \rangle / \tau_1)}]
% \end{align}

% \begin{equation}\label{equ2}
% \begin{split}
% L_{vtc} = -\frac{1}{J}\sum_{j=1}^{J}\bigg(&\log \left[\frac{\exp(\left \langle\hat{v}^c_j, \hat{d}_j\right \rangle / \tau_1)}{\sum_{l=1,l\ne j}^{J}\exp(\left \langle\hat{v}^c_j, \hat{d}_l\right \rangle / \tau_1)}\right] \\
% &+ \log \left[\frac{\exp(\left \langle\hat{v}^c_j, \hat{d}_j\right \rangle / \tau_1)}{\sum_{l=1,l\ne j}^{J}\exp(\left \langle\hat{v}^c_l, \hat{d}_j\right \rangle / \tau_1)}\right]\bigg)
% \end{split}
% \end{equation}

\begin{equation}\label{equ2}
\begin{split}
L_{vtc} = -\frac{1}{6}\sum_{j=1}^{6}\bigg(&\log \left[\frac{\exp(\left \langle\hat{v}^c_j, \hat{d}_j\right \rangle / \tau_1)}{\sum_{l=1,l\ne j}^{6}\exp(\left \langle\hat{v}^c_j, \hat{d}_l\right \rangle / \tau_1)}\right] \\
&+ \log \left[\frac{\exp(\left \langle\hat{v}^c_j, \hat{d}_j\right \rangle / \tau_1)}{\sum_{l=1,l\ne j}^{6}\exp(\left \langle\hat{v}^c_l, \hat{d}_j\right \rangle / \tau_1)}\right]\bigg)
\end{split}
\end{equation}
where $\left \langle\hat{v}^c_j, \hat{d}_j\right \rangle $ refers to the cosine similarity between the  embedding of the visual [CLS] token and the embedding of the textual [CLS] token.
% where $\left \langle\hat{v}^c_j, \hat{d}_j\right \rangle $ refers to the cosine similarity between the global image feature $\hat{v}^c_j$ of C-STMap $m^c_j$ and the global textual embedding $\hat{d}_j$ of text description $d_j$.
$\tau_1$ is the temperature to scale logits.
% $J$ represents the number of generated positive vision-text pairs.

\textbf{Frequency contrastive loss} ($ L_{fc}$): This loss aims to pull close rPPG signals ($ Y^p=\{y^p_1,y^p_2\} $) from positive samples while pushing them away from rPPG signals ($ Y^n=\{y^n_1,y^n_2,y^n_3\}$) from negative samples in the feature space. 
% $ L_{fc} $ can optimize the embedding space of the vision encoder to be discriminative to skin color changes. 
We write out $ L_{fc}$ as:
\begin{equation}\label{equ3}
L_{fc}=\log(\frac{\exp(e(y^p_1,y^p_2)/\tau_2 )}{ {\textstyle \sum_{i=1}^{3}} (\exp(e(y^p_1,y^n_i)/\tau_2 )+\exp(e(y^p_2,y^n_i)/\tau_2 ))}+1)
\end{equation}
where $e(y^p_1,y^p_2)$ represents the mean squared error between power spectral densities (PSD) of signal $y^p_1$ and $y^p_2$.

\textbf{Frequency ranking loss} ($L_{fr}$): Given that the negative samples $X^n$ are augmented from the given video based on a series of frequency ratios $R = \{r_1, r_2, r_3\}$, their predicted negative signals $Y^n$, along with the positive signals $Y^n$ from positive samples $X^p$, should conform to specific frequency ranks. Therefore, we propose $L_{fr}$ to optimize the ranks according to the main frequencies of signals $Y^p$ and $Y^n$. We achieve it via constraining the relative ranks of signal frequencies, offering a flexible way to avoid overfitting.
% the signal frequency ratios between the predicted positive and negative signals ($Y^p$ and $Y^n$) should be consistent with these ratios. \cite{yue_facial_2023} introduces the frequency ratio consistency loss $ L_{frc}$ with an absolute term to enhance the vision encoder's ability to predict signals with specific target frequencies. \ZJ{We instead propose $L_{fr}$ with a relative term to optimize the frequency ranks according to the main frequencies of signals $Y^p$ and $ Y^n $, offering a more flexible way to avoid overfitting caused by the strong constraints in the absolute term of $L_{frc}$.} 

Specifically, the ground truth frequency ranks for $Y = \{y^p_1,y^p_2,y^n_1,y^n_2,y^n_3\}$ are given as multi-level ratings $L = \{l(1), \ldots, l(5)\}$, where $l(q) \in \{1, \ldots, 4\}$ is determined by the ratios $ R $. It should be noticed that $l(1) = l(2)$ because $ y^p_1$ and $y^p_2$ share the same main frequency so they should be assigned with the same frequency rank. If $l(q) > l(c)$, it indicates that the signal of the former has a higher main frequency than that of the latter. For example, supposedly we leverage the sampled ratios $R = \{r_1=0.5, r_2=1.5, r_3=0.75\}$ to augment negative samples $X^n$, which contain rPPG signals with frequencies of $\{0.5 \times f^a, 1.5 \times f^a, 0.75 \times f^a\}$; the ground truth frequency ranks for $Y$ should be assigned as $L = \{3,3,1,4,2\}$. We denote the predicted frequency ranks for $Y$ as $S = \{s(1), \ldots, s(5)\}$ and define $L_{fr}$ as:
\begin{equation}\label{equ4}
L_{fr} =  \textstyle \frac{1}{9} \sum_{q=1}^{5}\sum_{c=1,l(c)<l(q)}^{5} \phi(s(q) - s(c))
\end{equation}
where $\phi$ is the logistic function $\phi(u) = \log(1+e^{-u})$. 
% \ZJ{$ L_{fr}$ enforces the predicted frequency ranks for videos containing rPPG signals of higher/lower main frequencies to be greater/lower than those from other videos.} 

% \ZJ{$L_{fr}$ enforces the vision encoder to predict that STMaps with relatively higher/lower rPPG frequencies in $M=\{M^p,M^n\}$ should exhibit correspondingly higher/lower frequency ranks. This mechanism improve its robustness to handle STMaps with diverse frequencies of rPPG signals.}
%\ZJ{$L_{fr}$ enforces the vision encoder to predict that the STMaps of samples in $X=\{X^p,X^n\}$, which have relatively higher/lower rPPG frequencies, should exhibit correspondingly higher/lower frequency ranks.
This loss strengthens the model's capability to effectively differentiate diverse frequencies of color variations among STMaps.

% enhances the model to distinguish facial videos with diverse frequencies of rPPG signals. This 

% predicted frequency ranks for videos containing rPPG signals of higher/lower main frequencies to be greater/lower than those from other videos.}

\medskip 

The overall loss function is the combination of the above three losses as well as the reconstruction loss in Sec.~\ref{sec3.5.3}:
\begin{equation}\label{equ5}
L= L_{r} + L_{vtc} + L_{fc} + L_{fr}
\end{equation}

\section{Experiments}

\subsection{Datasets}\label{sec4.1}

We conduct experiments using four benchmark datasets: UBFC-rPPG \cite{bobbia_unsupervised_2019}, PURE \cite{stricker_non-contact_2014}, VIPL-HR \cite{niu_rhythmnet_2020} and MMVS \cite{yue_deep_2021}.

UBFC-rPPG consists of 42 facial videos recorded from 42 subjects using a Logitech C920 HD Pro webcam with a resolution of $640 \times 480$ and with 30 FPS. We follow the protocol outlined in \cite{lu_dual-gan_2021} to spilt the training and test sets.

PURE consists of 60 one-minute facial videos from 10 subjects with six kinds of activities, including stable, talking, slow translation, fast translation, small head rotation, and medium head rotation. The videos were captured at a frame rate of 30 FPS and a resolution of $640 \times 480$. The ground truth PPG signals were recorded using a finger clip pulse oximeter. To ensure consistency, we follow the same experimental protocol used in \cite{lu_dual-gan_2021} to split the training and test sets.

VIPL-HR contains 2,378 facial videos from 107 subjects. These videos are collected by three different devices (\ie, Logitech C310, HUAWEI P9, and RealSense F200) in nine less-constrained scenarios (\ie, stable, motion, talking, dark, bright, long distance, exercise, phone stable, and phone motion scenarios). We follow the protocol in \cite{yu_physformer_2022} to split the training and test sets.

MMVS contains 745 facial videos from 129 subjects, with 560 videos recorded in a laboratory setting and 185 videos collected in a hospital environment. A depth camera Intel Realsense D435i was utilized to record facial videos with the resolution of $1920 \times 1080$ and the frame rate of 25 FPS. A finger clip pulse oximeter Contec CMS50E was used to record the ground truth PPG signals with the sampling rate of 60 Hz. We follow \cite{yue_facial_2023} for the train-test split.

\subsection{Implementation Details}\label{sec4.2}

\indent \textbf{Vision-Language model.} We employ the vision and text encoders from the widely-used VindLU \cite{cheng_vindlu_2023}, specifically the ViT-B/16 and BERT pre-trained across large-scale video-text and image-text datasets, \ie CC3M \cite{sharma-etal-2018-conceptual}, CC12M \cite{sharma-etal-2018-conceptual}, Visual Genome \cite{krishna_visual_2017}, SBU captions \cite{ordonez_im2text_2011}, and WebVid10M \cite{bain_2021}. 
Other choices of VLMs can be found in our ablation study. 
%Notice those famous VLMs such as CLIP, BLIP, or COCA  were not suitable here as they were only pre-trained on image-text datasets. 
We initialize the weights of the vision and text encoders with pre-trained parameters from VindLU, and further fine-tune them in context of our proposed VL-phys.

%  Conceptual Captions \cite{changpinyo_conceptual_2021}, Conceptual 12M \cite{changpinyo_conceptual_2021}, , LAION \cite{schuhmann_laion-400m_nodate}). 

\textbf{Hyper-parameters.} For facial videos in four datasets, we randomly sample consecutive $F=224$ frames from each video to build the spatio-temporal maps (STMaps) $M^p=\left \{ m^p \right \}$ and $M^n=\left \{ m^n \right \}$ for positive and negative samples, respectively. {We follow} \cite{niu_rhythmnet_2020} {to set the number of detected ROIs in facial videos as $A=25$.} The dimension of the STMap is $224\times 224\times 3$, ensuring compatibility with the size requirement for the ViT-based vision encoder. 
% Additionally, a max-min normalization is performed on each STMap as stated in \cite{qian_dual-path_2024}, uniformly scaling the pixel values into a range of [0, 255]. 
The temperature $\tau_1$ in equation (2) is set as 0.07 following \cite{radford_learning_2021}, and $\tau_2$ in equation (3) is set as 0.08 following \cite{yue_facial_2023}. The masking ratio $b=60\%$ is equivalent to that in \cite{kwon_masked_2023}. We freeze the weights of the pre-trained LFA module.

\textbf{Training.}  The training epoch is set at 100 for UBFC-rPPG, PURE and MMVS datasets, and extended to 200 for the large-scale dataset VIPL-HR. Our model is trained using the AdamW optimizer \cite{loshchilov_decoupled_2019} with a learning rate of $5\times10^{-5}$ and a batch size of 16. The training is conducted on four NVIDIA A40 GPUs.

\subsection{Evaluation Protocol}\label{sec4.3}

Previous works calculate the heart rate (HR), heart rate variability (HRV) and respiration frequency (RF) from the estimated rPPG signals and compare them to the ground truth PPG signals for performance evaluation \cite{yu_physformer_2022,sun_contrast-phys_2024,liu_rppg-mae_2024}. We follow them to conduct intra-dataset HR evaluation on four datasets; HRV and RF evaluation on the UBFC-rPPG dataset. Moreover, we perform cross-dataset HR evaluation among UBFC-rPPG, PURE and MMVS. The calculation of HR, HRV and RF is via the toolkit HeartPy \cite{van_gent_analysing_2019}.

In line with prior research \cite{yu_physformer_2022,sun_contrast-phys_2024,liu_rppg-mae_2024}, we use three metrics to assess the accuracy for HR evaluation: the mean absolute error (MAE), root mean squared error (RMSE), and Pearson correlation coefficient ($\rho$). For the evaluation of HRV features, which encompasses low-frequency power (LF) in normalized units (n.u.), highfrequency power (HF) in normalized units (n.u.), and the LF/HF power ratio, we report the standard deviation (Std) of estimation errors, RMSE, and $\rho$ as evaluation metrics. Finally, for RF, we also report the Std, RMSE and $\rho$ as per most comparable methods \cite{yu_physformer_2022,sun_contrast-phys_2024,liu_rppg-mae_2024}.

\begin{table*}[!t]
\caption{Comparison to state of the art on HR estimation. The results are reported on UBFC-rPPG, PURE, VIPL-HR, and MMVS datasets. $\uparrow$ indicates that the larger the value is the better it is and $\downarrow$ vice versa. The best supervised approach is marked in \colorbox{gray!50}{shadow}, while the best self-supervised approach is marked in \textbf{bold}.}
\vspace{-0.1in}
\label{table1}
\begin{center}
\setlength{\tabcolsep}{0.7mm}
\hspace*{-1cm} % 负值表示向左移动，调整数值以达到合适的位置
\footnotesize % 调整表格的字号为小一号
\begin{tabular}{c|c|ccc|ccc|ccc|ccc}
\toprule
\multirow{2}*{Method} &PPG &\multicolumn{3}{c|}{UBFC-rPPG}& \multicolumn{3}{c|}{PURE}& \multicolumn{3}{c|}{VIPL-HR}& \multicolumn{3}{c}{MMVS} \\ 
& annotations&MAE$\downarrow$  &RMSE$\downarrow$  &$\rho\uparrow$  & MAE$\downarrow$  &RMSE$\downarrow$  &$\rho\uparrow$ & MAE$\downarrow$  &RMSE$\downarrow$  &$\rho\uparrow$ & MAE$\downarrow$  &RMSE$\downarrow$  &$\rho\uparrow$ \\ 
\midrule
POS &- &8.35  &10.00  &0.24  &3.14  &10.57  &0.95  &11.53  &17.26  &0.30 &6.77 &9.40 &0.82 \\
CHROM &- &8.20  &9.92  &0.27  &3.82  &6.80  &0.97  &11.42  &16.92  &0.28 &6.85 &9.37  &0.82 \\
Green  &- &6.01  &7.87  &0.29  &4.39  &11.60  &0.90  &-  &-  &- &7.13 &9.46  &0.80 \\
SynRhythm  &\ding{51}  &5.59  &6.82  &0.72  &2.71  &4.86 &0.98  &-  &-  &- &4.48 &6.52  &0.89 \\
Meta-rppg &\ding{51} &5.97  &7.42  &0.53  &2.52  &4.63 &0.98  &-  &-  &- &4.30 &6.20  &0.91 \\
PulseGan &\ding{51} &1.19 &2.10  &0.98  &2.28  &4.29 &0.99  &-  &-  &- &3.52 &5.09  &0.93 \\ 
Dual-Gan &\ding{51} &0.44  &0.67  &0.99  &0.82 &1.31  &0.99 &\colorbox{gray!50}{4.93}  &\colorbox{gray!50}{7.68}  &\colorbox{gray!50}{0.81} &\colorbox{gray!50}{3.00} &\colorbox{gray!50}{4.27} &\colorbox{gray!50}{0.94} \\
Physformer &\ding{51} &0.40 &0.71  &0.99  &1.10  &1.75 &0.99  &4.97 &7.79  &0.78 &3.28 &4.50  &0.93 \\ 
Du \etal&\ding{51}  &\colorbox{gray!50}{0.16} &0.57  &0.99  &-  &-  &-   &-  &-  &- &-  &-  &-\\ 
Dual-TL&\ding{51}  &0.17 &\colorbox{gray!50}{0.41}  &\colorbox{gray!50}{0.99}  &0.37  &0.68  &0.99   &6.38  &6.92  &0.69 &-  &-  &-\\ 
ND-DeeprPPG &\ding{51}  &0.31 &0.98  &0.99  &\colorbox{gray!50}{0.18}  &\colorbox{gray!50}{0.41}  &\colorbox{gray!50}{0.99}  &-  &-  &- &-  &-  &-\\ 
\midrule
Gideon \etal  &\ding{55} &1.85  &4.28  &0.93  &2.32  &2.97  &0.99  &9.80  &15.48  &0.38 &3.43 &4.74 &0.93 \\
Yue \etal &\ding{55} &0.58  &0.94 &0.99 &1.23 &2.01  &0.99  &-  &- &- &2.93 &4.16  &0.94 \\ 
Contrast-Phys+ &\ding{55} &0.64  &1.00 &0.99 &1.00 &1.40  &0.99  &7.49  &14.40 &0.49 &2.96 &4.11  &0.94 \\ 
SiNC &\ding{55} &0.59  &1.83 &0.99 &0.61 &1.84  &0.99  &-  &- &- &3.15 &4.40  &0.93 \\
VL-phys &\ding{55} &\textbf{0.28}  &\textbf{0.60} &\textbf{0.99} &\textbf{0.52} &\textbf{0.95}  &\textbf{0.99}  &\textbf{6.04}  &\textbf{8.78} &\textbf{0.71} &\textbf{2.33} &\textbf{3.76}  &\textbf{0.95} \\ 

\bottomrule
\end{tabular}
\end{center}
\end{table*}

\begin{figure}[t]
\begin{center}
\begin{tabular}{cc}
\includegraphics[width=1.75in]{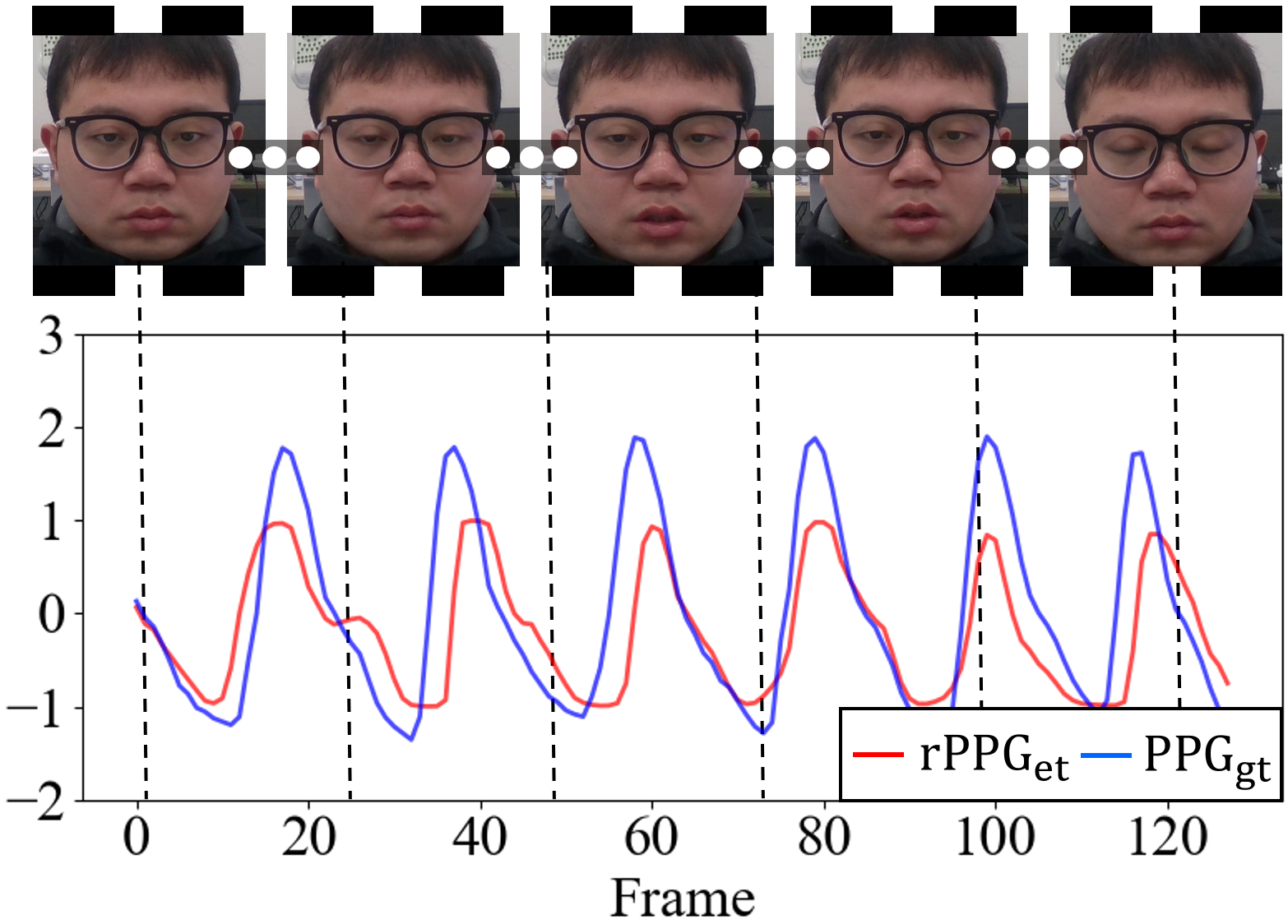} &
\includegraphics[width=1.75in]{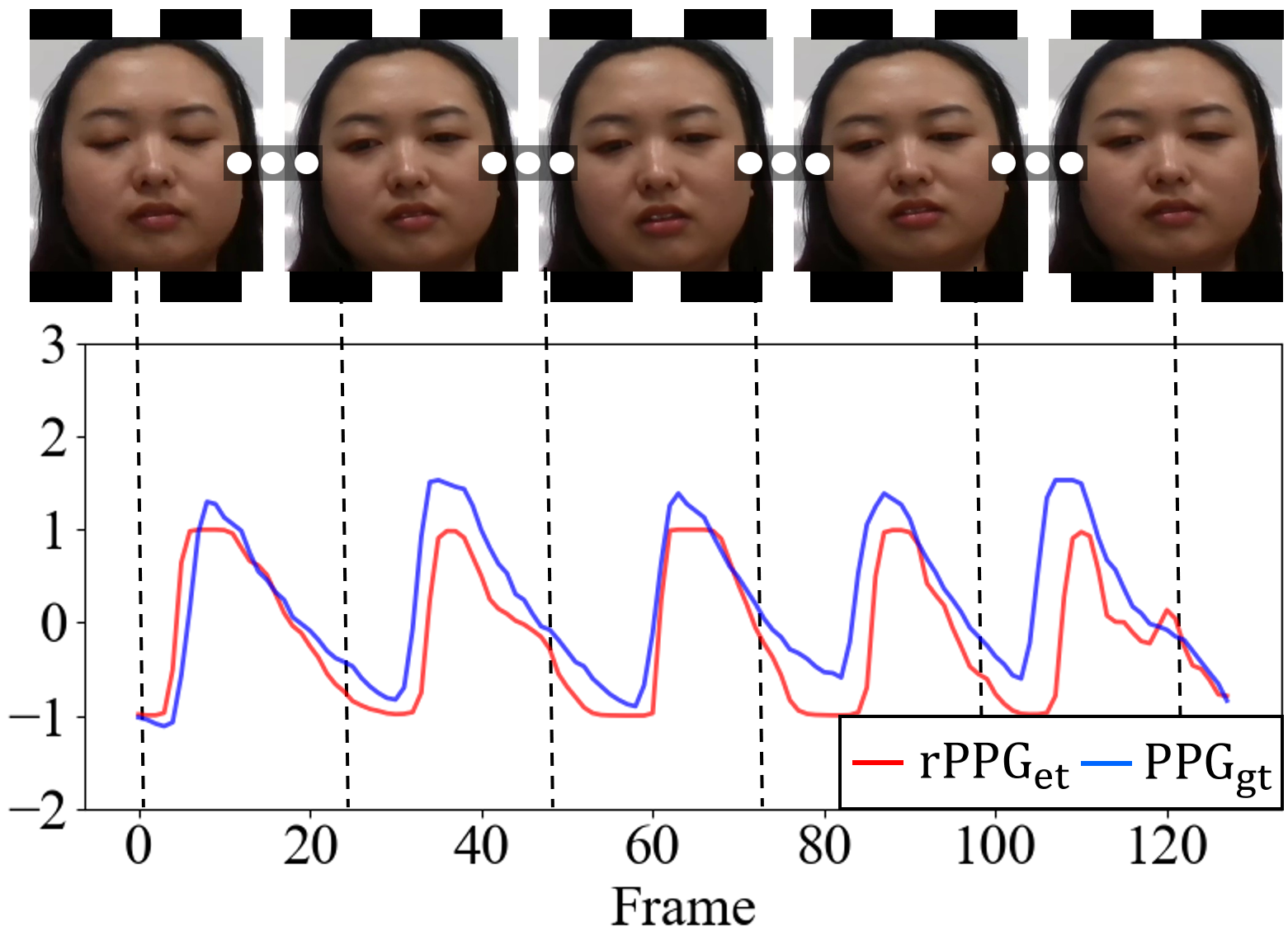} \\
(a) & (b) \\
\includegraphics[width=1.75in]{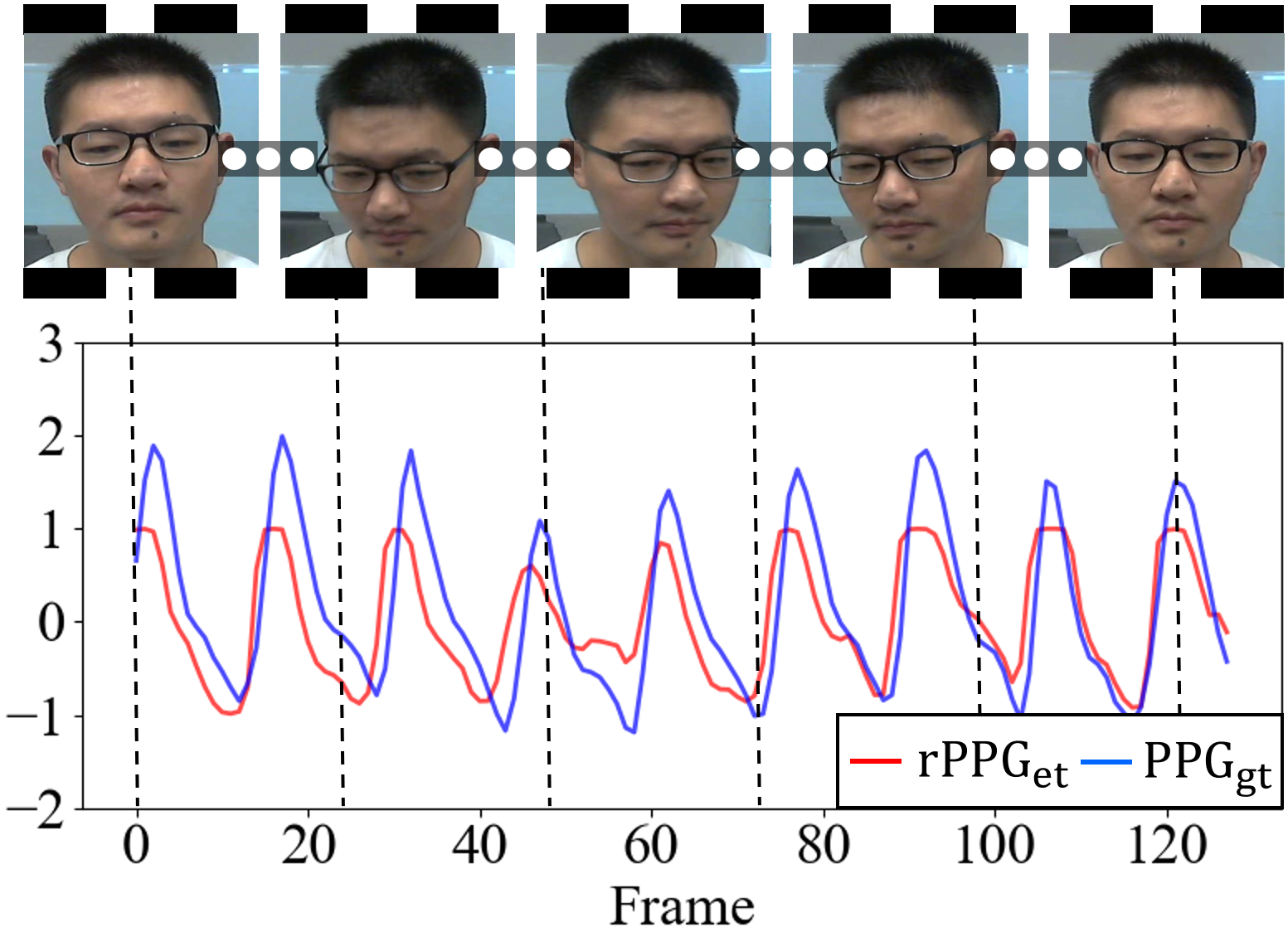} &
\includegraphics[width=1.75in]{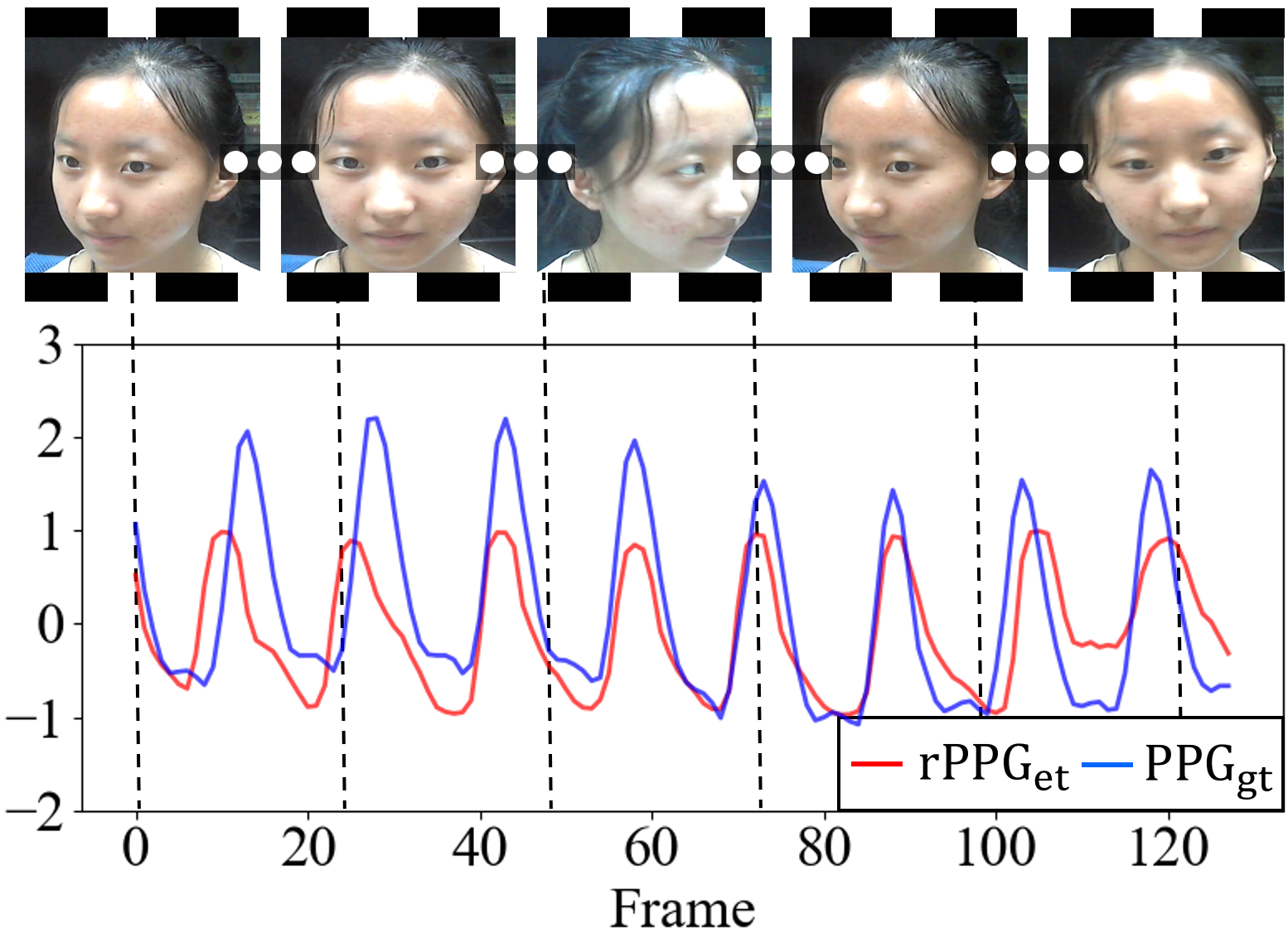} \\
(c) & (d) \\
\includegraphics[width=1.75in]{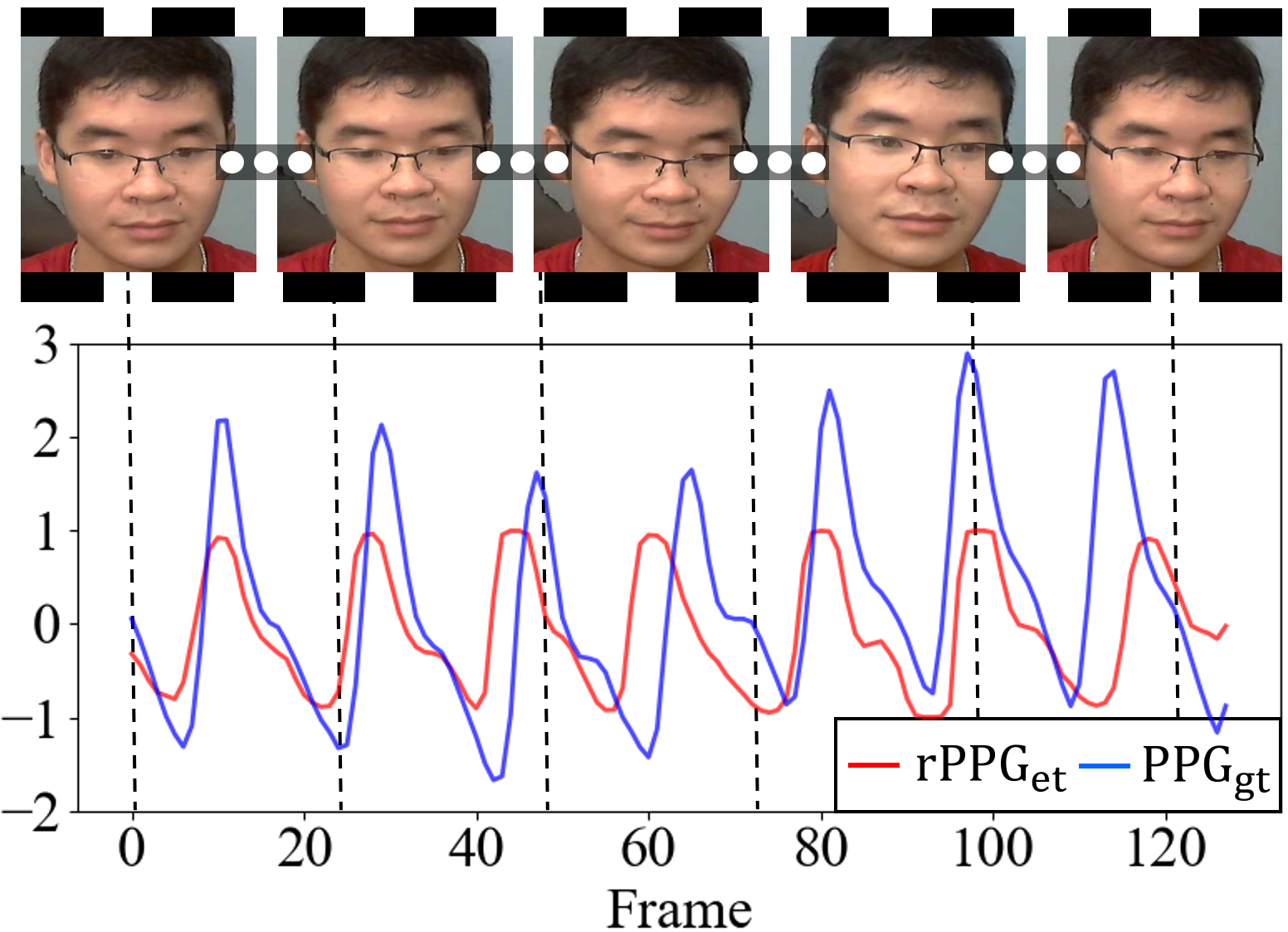} &
\includegraphics[width=1.75in]{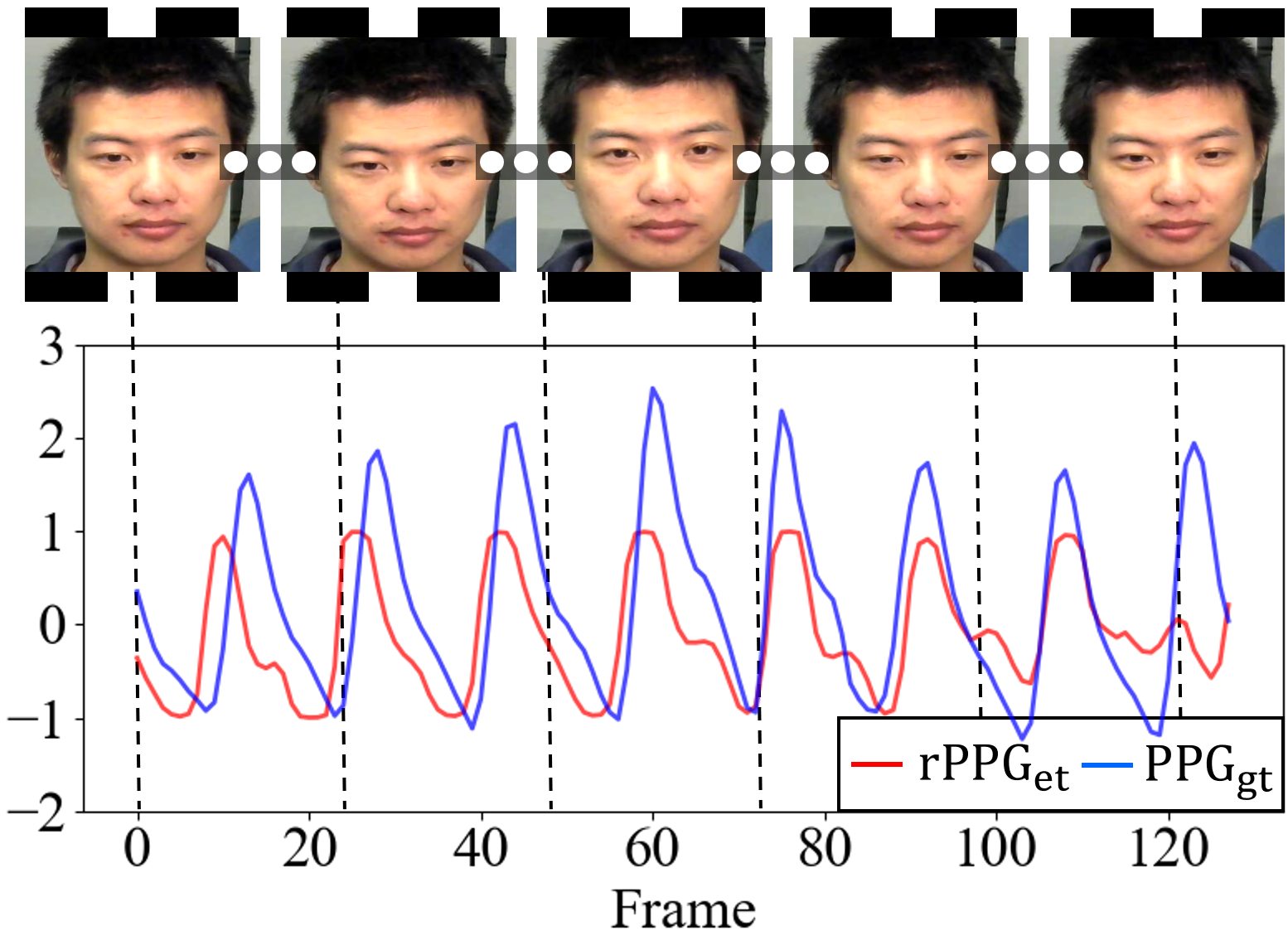} \\
(e) & (f) \\
\end{tabular}
\end{center}
\caption{Six examples for the visual comparison between estimated rPPG signals (red curves) and their corresponding ground truth PPG signals (blue curves).}
\label{fig5}
\end{figure}

\begin{figure}[t]
\begin{center}
\begin{tabular}{cc}
\includegraphics[width=1.75in]{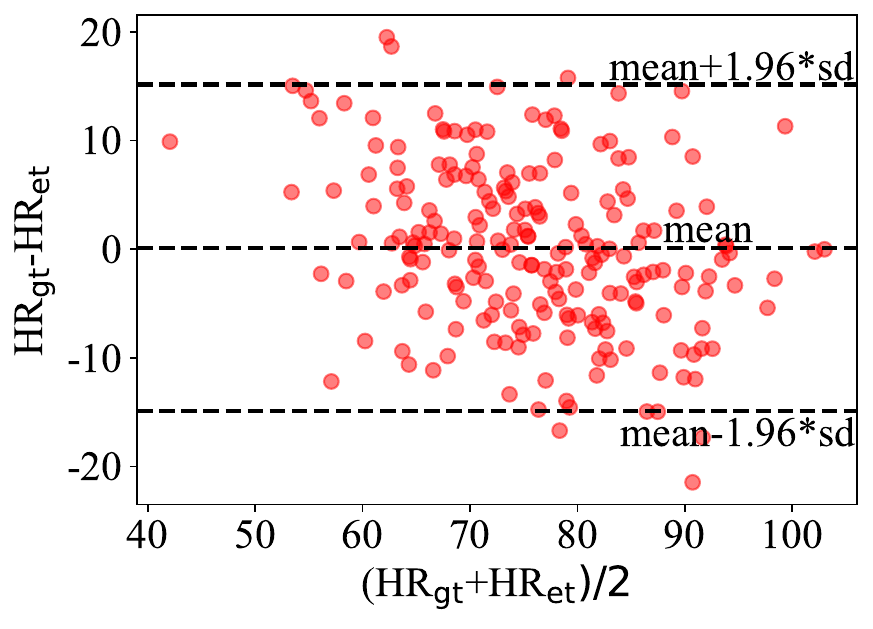} &
\includegraphics[width=1.75in]{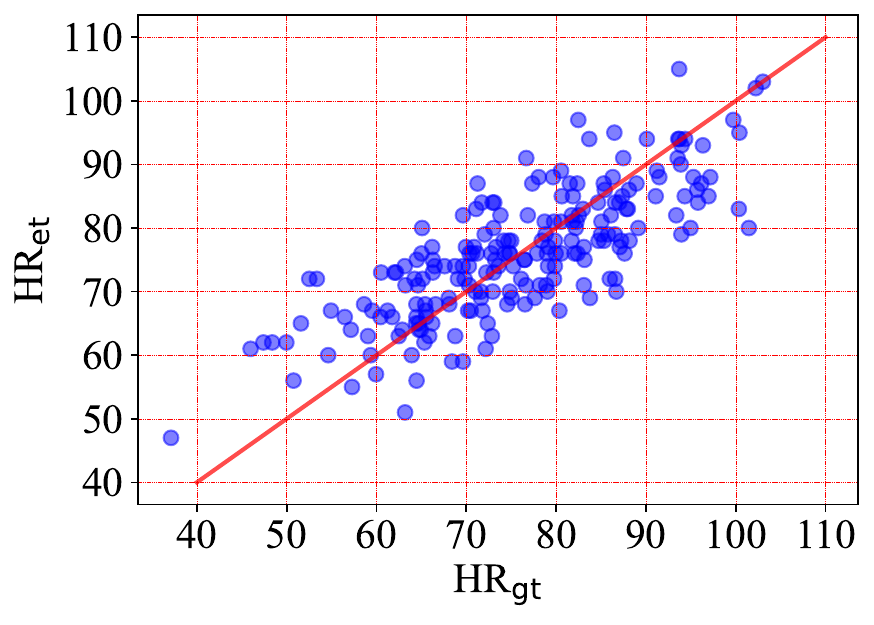} \\
(a) VIPL-HR & (b) VIPL-HR\\
\includegraphics[width=1.75in]{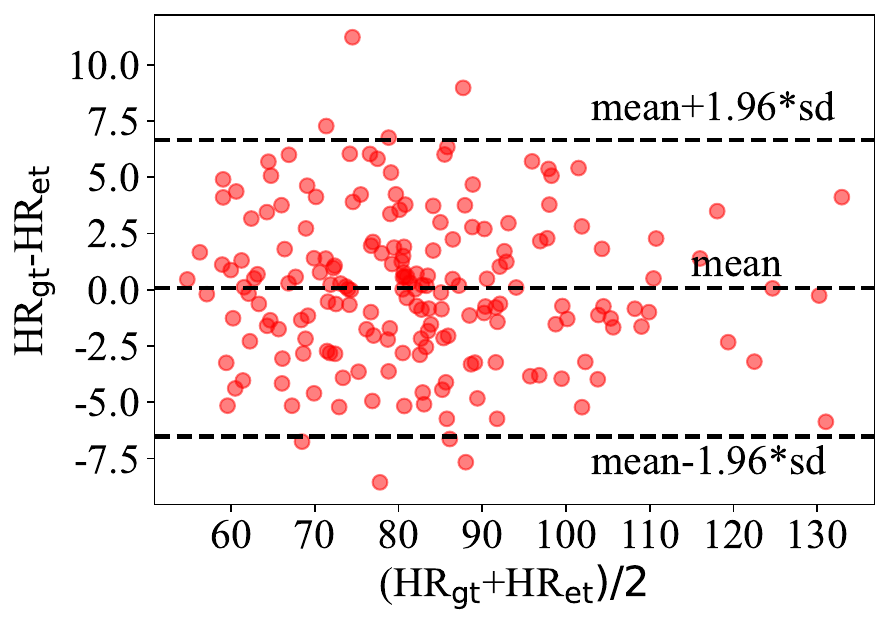} &
\includegraphics[width=1.75in]{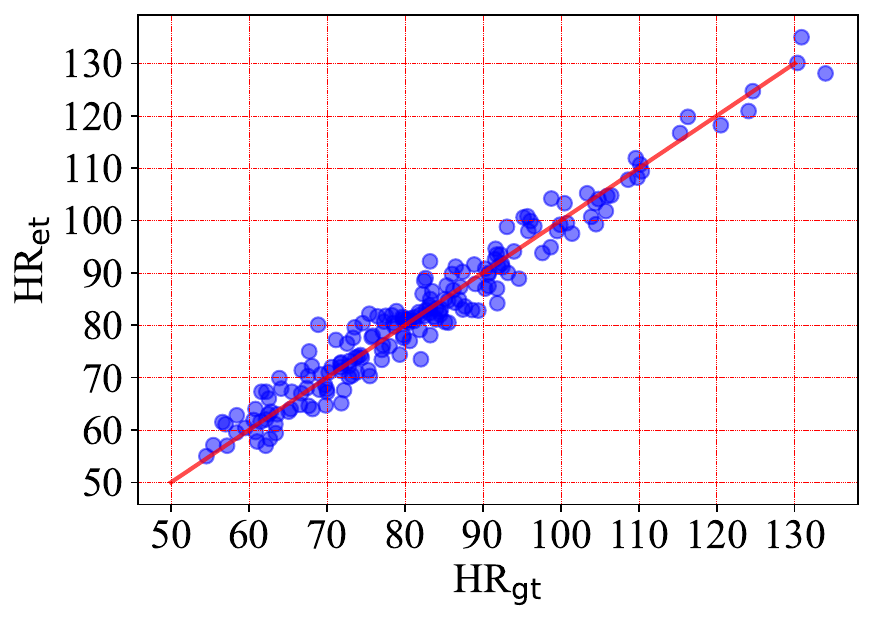} \\
(c) MMVS& (d) MMVS\\

\end{tabular}
\end{center}
\caption{The Bland-Altman plots and scatter plots show the difference between estimated HR and ground truth HR on the VIPL-HR and MMVS datasets.}

\label{fig6}
\end{figure}

\subsection{Results}
\subsubsection{HR evaluation}

We first conduct intra-dataset HR estimation on four datasets. As shown in Table \ref{table1}, we compare the proposed VL-phys with three traditional methods (POS \cite{wang_algorithmic_2017}, CHROM \cite{de_haan_robust_2013} and Green \cite{verkruysse_remote_2008}); eight DNN-based supervised methods (SynRhythm \cite{niu_synrhythm_2018}, Meta-rppg \cite{lee_meta-rppg_2020}, PulseGan \cite{song_pulsegan_2021}, Dual-Gan \cite{lu_dual-gan_2021}, Physformer \cite{yu_physformer_2022}, Du \etal \cite{du_dual-bridging_2023}, Dual-TL \cite{qian_dual-path_2024}, and ND-DeeprPPG \cite{liu_robust_2024}); four DNN-based self-supervised methods (Gideon \etal \cite{gideon_way_2021}, Contrast-Phys+ \cite{sun_contrast-phys_2024}, Yue \etal \cite{yue_facial_2023}, and SiNC \cite{speth_non-contrastive_2023}). 

First, it is evident that traditional methods exhibit poor performance across four datasets. For example, the metric $\rho$ for POS \cite{wang_algorithmic_2017} and CHROM \cite{de_haan_robust_2013} on UBFC-rPPG dataset is lower than 0.3. Traditional methods typically rely on the assumption that target rPPG signals can be separated from the RGB space by linear projection. However, this assumption proves overly idealistic, as periodic rPPG signals are easily intertwined with non-periodic noises in facial videos, which makes them hard to be disentangled.

Second, 
% the performance gap between DNN-based supervised approaches and self-supervised approaches is relatively narrow. 
%DNN-based self-supervised approaches show fairly good performance to supervised ones. 
the performance gap between VL-phys and DNN-based supervised ones is narrow.
% Our approach outperforms many supervised approaches. 
For instance, on the UBFC-rPPG dataset, it is only inferior to two advanced models Du \etal \cite{du_dual-bridging_2023} and Dual-TL \cite{qian_dual-path_2024}. On the most challenging dataset VIPL-HR, we achieve the MAE of 6.04 and the RMSE of 8.78, which are also comparable to the best-performing one Dual-Gan \cite{lu_dual-gan_2021}. Surprisingly, on the MMVS dataset, it even decreases the MAE/RMSE from Dual-Gan by 0.67/0.51.

Third, our VL-phys achieves the best performance among self-supervised methods across all metrics. For example, on UBFC-rPPG, it decreases MAE by 0.30 and RMSE by 0.34 from Yue \etal \cite{yue_facial_2023}. On PURE, it decreases 0.48/0.45 on MAE/RMSE from the very recent work Contrast-Phys+ \cite{sun_contrast-phys_2024}.

\{Moreover, it is worth noting that the general performance of comparable methods on the VIPL-HR dataset is clearly inferior to that on the rest datasets. This performance discrepancy arises because the facial videos in the VIPL-HR dataset were collected in less-constrained scenarios, including diverse ambient lighting conditions, and varying camera angles. These challenging factors introduce noises and hinder the capturing of physiological clues for accurate rPPG estimation. In contrast, other datasets were collected under controlled conditions, with consistent lighting and stable camera positioning.}

Besides the quantitative results, we also visualize the results of rPPG predictions in Fig.~\ref{fig5}. We can observe that the predicted rPPG signals are close to the ground truth PPG signals in both frequency and amplitude. In particular, for videos with special cases, such as subjects having head movements (Fig.~\ref{fig5}(c) and Fig.~\ref{fig5}(d)), bright (Fig.~\ref{fig5}(e)) and dim illumination (Fig.~\ref{fig5}(f)), the periodic patterns of peaks and troughs of the predicted rPPG signals are still consistent with the ground truth. This demonstrates the generalizability of the proposed method.

Next, we show the scatter plots and Bland-Altman plots for VIPL-HR and MMVS datasets in Fig.~\ref{fig6}. $HR_{gt}$ and $HR_{et}$ represent the HR calculated from the ground truth PPG signal and the estimated rPPG signal, respectively. The top and bottom dashed lines indicate confidence intervals for 95\% limits of agreement. We can observe that, for the VIPL-HR dataset (Fig.~\ref{fig6}(a)), the confidence interval ranges from -14.30 to 14.83 BPM (beats per minute). For the MMVS dataset (Fig.~\ref{fig6}(c)), the interval is from –6.98 to 7.29 BPM. These intervals confirm that our results have a strong correlation with the ground truth. In addition, from Fig.~\ref{fig6}(b) and Fig.~\ref{fig6}(d) we observe that the estimated HR 
%\st{is close to the linear relationship with the ground truth HR and} 
closely approximates a linear relationship with the ground truth HR, 
% with most points 
% lie close to 
% situated near the fitting line,} 
indicating the effectiveness and robustness of the proposed method.

\subsubsection{RF and HRV evaluation}

We further conduct intra-dataset RF and HRV estimation on UBFC-rPPG. We compare our approach with state of the art, and present the results in Table \ref{table2}. LF, HF, and LF/HF are three attributes for HRV estimation. We can observe that the proposed method significantly outperforms traditional methods and most DNN-based methods. These findings underscore the capability of our method to yield high-quality rPPG signals with accurate systolic peaks, resulting into superior HRV estimations.

\begin{table*}[!t]
\caption{Comparison to state of the art on RF and HRV estimation. The results are reported on the UBFC-rPPG. $\uparrow$ indicates that the larger the value is the better {it is} and $\downarrow$ vice versa. The best supervised approach is marked in \colorbox{gray!50}{shadow}, while the best self-supervised approach is  marked in \textbf{bold}.}
\vspace{-0.1in}
\label{table2}
\begin{center}
\setlength{\tabcolsep}{1mm}
\hspace*{-2.2cm} % 负值表示向左移动，调整数值以达到合适的位置
\footnotesize % 调整表格的字号为小一号
\begin{tabular}{c|c|ccc|ccc|ccc|ccc}
\toprule
% &&&&&\multicolumn{9}{c}{} \\ 
\multirow{2}*{Method} &PPG&\multicolumn{3}{c|}{RF}& \multicolumn{3}{c|}{HRV: LF}& \multicolumn{3}{c|}{HRV: HF}& \multicolumn{3}{c}{HRV: LF/HF}\\

& annotations&Std$\downarrow$  &RMSE$\downarrow$  &$\rho\uparrow$  & Std$\downarrow$  &RMSE$\downarrow$  &$\rho\uparrow$ & Std$\downarrow$  &RMSE$\downarrow$  &$\rho\uparrow$ & Std$\downarrow$  &RMSE$\downarrow$  &$\rho\uparrow$\\ 
\midrule
POS &- &0.109  &0.107  &0.087  &0.171  &0.169  &0.479  &0.171  &0.169  &0.479 &0.405 &0.399 &0.518\\
CHROM&-&0.086  &0.089  &0.102  &0.243  &0.240  &0.159 &0.243  &0.240  &0.159 &0.655 &0.645  &0.266\\
Green&-&0.087  &0.086  &0.111  &0.186  &0.186  &0.280  &0.186  &0.186  &0.280 &0.361 &0.365  &0.492\\
CVD &\ding{51} &0.017  &0.018  &0.252  &0.053  &0.065 &0.740  &0.053  &0.065  &0.740 &0.169 &0.168  &0.812\\
rPPGNet &\ding{51} &0.030  &0.034  &0.233  &0.071 &0.070 &0.686  &0.071  &0.070 &0.686 &0.212 &0.208  &0.744\\
Dual-Gan &\ding{51} &0.010  &0.010  &0.395  &0.034 &0.035  &0.891  &0.034  &0.035  &0.891 &0.131 &0.136  &0.881\\
Physformer &\ding{51} &\colorbox{gray!50}{0.009}  &\colorbox{gray!50}{0.009}  &\colorbox{gray!50}{0.413}  &\colorbox{gray!50}{0.030} &\colorbox{gray!50}{0.032} &\colorbox{gray!50}{0.895}  &\colorbox{gray!50}{0.030} &\colorbox{gray!50}{0.032} &\colorbox{gray!50}{0.895} &\colorbox{gray!50}{0.126} &\colorbox{gray!50}{0.130}  &\colorbox{gray!50}{0.893}\\
\midrule
Gideon \etal &\ding{55} &0.061  &0.098  &0.103  &0.091  &0.139  &0.694  &0.091  &0.139  &0.694  &0.525 &0.691  &0.684\\
Yue \etal &\ding{55} &0.023  &0.028  &0.351  &0.047  &0.062  &0.769  &0.047  &0.062  &0.769 &0.160 &0.164  &\textbf{0.831}\\ 
Contrast-Phys+  &\ding{55} &0.085  &0.083  &0.347  &0.096  &0.098  &0.798  &0.096  &0.098  &0.798 &0.391 &0.395  &0.782\\ 
VL-phys &\ding{55} &\textbf{0.016}  &\textbf{0.019}  &\textbf{0.378}  &\textbf{0.039} &\textbf{0.046}  &\textbf{0.844}  &\textbf{0.039}  &\textbf{0.046}  &\textbf{0.844} &\textbf{0.142} &\textbf{0.147} &\textbf{0.868}\\ 

\bottomrule
\end{tabular}
\end{center}
\end{table*}

\begin{table*}[!t]
\caption{Comparison to state of the art on cross-dataset HR estimation.  $\uparrow$ indicates that the larger the value is the better {it is} and $\downarrow$ vice versa. The best supervised approach is marked in \colorbox{gray!50}{shadow}, while the best self-supervised approach is  marked in \textbf{bold}.}
\vspace{-0.1in}
\label{table3}
\begin{center}
\setlength{\tabcolsep}{1.5mm}
\hspace*{-2.1cm} % 负值表示向左移动，调整数值以达到合适的位置
\footnotesize % 调整表格的字号为小一号
\begin{tabular}{c|c|ccc|ccc|ccc|ccc}
\toprule
\multirow{2}*{Method}  &PPG &\multicolumn{3}{c|}{MMVS$\rightarrow$UBFC-rPPG} &\multicolumn{3}{c|}{UBFC-rPPG$\rightarrow$MMVS}&\multicolumn{3}{c|}{PURE$\rightarrow$UBFC-rPPG}&\multicolumn{3}{c}{UBFC-rPPG$\rightarrow$PURE}\\ 
&annotations  &MAE$\downarrow$  &RMSE$\downarrow$  &$\rho\uparrow$  & MAE$\downarrow$  &RMSE$\downarrow$  &$\rho\uparrow$ & MAE$\downarrow$  &RMSE$\downarrow$  &$\rho\uparrow$ & MAE$\downarrow$  &RMSE$\downarrow$  &$\rho\uparrow$\\ 
\midrule
Meta-rppg  &\ding{51}&6.48 &7.97 &0.52 &5.69  &7.74  &0.84 &6.11 &7.58 &0.66 &4.00 &5.98 &0.92\\
PulseGan  &\ding{51}&2.33 &3.62 &0.97 &4.40  &6.35  &0.89  &2.30 &3.50 &0.97 &3.36 &5.11 &0.95 \\
Dual-Gan&\ding{51}&2.00 &3.13 &0.97 &3.51  &4.99  &0.93 &2.03 &3.01 &0.97 &1.81 &2.97 &0.99 \\
% RhythmNet \cite{niu_rhythmnet_2020} &$\surd$&4.02 &5.35 &0.86 &4.82  &6.63  &0.87 &3.65 &4.44 &0.90 &3.72 &5.75 &0.95 \\
Physformer &\ding{51} &\colorbox{gray!50}{1.97} &\colorbox{gray!50}{3.08} &\colorbox{gray!50}{0.97} &\colorbox{gray!50}{3.46}  &\colorbox{gray!50}{4.96}  &\colorbox{gray!50}{0.93} &1.93 &3.02 &0.97 &1.99 &3.28 &0.99 \\
ND-DeeprPPG &\ding{51}&- &- &- &-  &-  &-  &\colorbox{gray!50}{0.34} &\colorbox{gray!50}{0.98} &\colorbox{gray!50}{0.99} &\colorbox{gray!50}{0.17} &\colorbox{gray!50}{0.35} &\colorbox{gray!50}{0.99} \\
\midrule
Gideon \etal &\ding{55}&2.45 &3.71 &0.93 &3.84  &5.31  &0.91 &2.37 &3.51 &0.95 &2.95 &4.60 &0.97\\
Yue \etal &\ding{55}&2.24 &3.40 &0.97 &3.50  &4.95  &0.93 &2.18 &3.20 &0.97 &2.14 &3.37 &0.98\\ 
Contrast-Phys+  &\ding{55} &2.18 &3.42 &0.97 &3.55 &5.03 &0.92 &2.07 &3.10 &0.97 &2.03 &3.35 &0.99\\
VL-phys  &\ding{55} &\textbf{1.68} &\textbf{2.83} &\textbf{0.98} &\textbf{2.70} &\textbf{4.06} &\textbf{0.94} &\textbf{1.73} &\textbf{2.94} &\textbf{0.97} &\textbf{1.80} &\textbf{3.06} &\textbf{0.99}\\
\bottomrule
\end{tabular}
\end{center}
\end{table*}

\section{Discussion}

\subsection{Cross-dataset HR evaluation}\label{tab:cross-dataset}

To evaluate the generalization of our method, we perform cross-dataset HR evaluation among UBFC-rPPG, PURE and MMVS datasets. Specifically, we train VL-phys as well as recent supervised and self-supervised methods on one dataset and test them on another. For example, MMVS$\rightarrow$UBFC-rPPG means training on MMVS while testing on UBFC-rPPG. The results are presented in Table \ref{table3}. It is evident that VL-phys consistently achieves more robust results compared to other methods. For example, in the cross-dataset setting of MMVS$\rightarrow$UBFC-rPPG, VL-phys significantly reduces MAE from 2.18 to 1.68 compared to the latest self-supervised method, Contrast-Phys+ \cite{sun_contrast-phys_2024}. These results show the strong generalization ability of our method in the unknown scenario.

\begin{table*}[!t]
\caption{HR estimation of VL-phys on four datasets under semi-supervised and fully-supervised settings.}
\label{table4}
\begin{center}
\setlength{\tabcolsep}{1.5mm}
\hspace*{-1.5cm} % 负值表示向左移动，调整数值以达到合适的位置
\footnotesize % 调整表格的字号为小一号
\begin{tabular}{c|ccc|ccc|ccc|ccc}
\toprule
\multirow{2}*{Method} &\multicolumn{3}{c|}{UBFC-rPPG}& \multicolumn{3}{c|}{PURE}& \multicolumn{3}{c|}{VIPL-HR}& \multicolumn{3}{c}{MMVS} \\ 
&MAE$\downarrow$  &RMSE$\downarrow$  &$\rho\uparrow$  & MAE$\downarrow$  &RMSE$\downarrow$  &$\rho\uparrow$ & MAE$\downarrow$  &RMSE$\downarrow$  &$\rho\uparrow$ & MAE$\downarrow$  &RMSE$\downarrow$  &$\rho\uparrow$ \\ 
\midrule

% Dual-TL \cite{qian_dual-path_2023} &0.17 &0.41 &0.99 &- &- &-\\
% Dual-Gan \cite{lu_dual-gan_2021} &0.44 &0.67 &0.99 &3.00 &4.27 &0.94\\
VL-phys &0.28 &0.60 &0.99 &0.52 &0.95  &0.99  &6.04  &8.78 &0.71  &2.33 &3.76 &0.95\\
VL-phys(90\%+10\%) &0.14 &0.32 &0.99 &0.26 &0.63 &0.99 &5.32 &8.06 &0.76 &2.08 &3.35 &0.97\\
VL-phys(80\%+20\%) &0.13 &0.29 &0.99 &0.20 &0.47 &0.99 &4.90 &7.70 &0.80 &1.94 &3.00 &0.97\\
VL-phys(70\%+30\%) &0.13 &0.31 &0.99 &0.16 &0.44 &0.99 &4.86 &7.54 &0.82 &1.93 &{2.96} &{0.97}\\
{VL-phys(60\%+40\%)} &{0.12} &{0.30} &{0.99} &{0.16} &{0.44} &{0.99} &{4.86} &{7.55} &{0.82} &{1.93} &{2.96} &{0.97}\\
{VL-phys(50\%+50\%)} &{0.12} &{0.28} &{0.99} &{0.15} &{0.42} &{0.99} &{4.85} &{7.54} &{0.82} &{1.92} &{2.95} &{0.97}\\
{VL-phys(0\%+100\%)} &{\textbf{0.11}} &{\textbf{0.27}} &{\textbf{0.99}} &{\textbf{0.13}} &{\textbf{0.40}} &{\textbf{0.99}} &{\textbf{4.83}} &{\textbf{7.51}} &{\textbf{0.83}} &{\textbf{1.88}} &{\textbf{2.91}} &{\textbf{0.97}}\\

\bottomrule
\end{tabular}
\end{center}
\end{table*}

\subsection{Semi-supervised and fully-supervised HR Evaluation}

{We then evaluate the performance of VL-phys on four datasets under the semi-supervised setting and the fully-supervised setting,} as shown in Table \ref{table4}. For instance, VL-phys (90\%+10\%) indicates that 90\% samples in the training set are utilized for self-supervised learning, while the remaining 10\% are used for supervised learning. 
% This is achieved by incorporating another supervision between the predicted rPPG signals with the ground truth PPG signals. 
The two types of data are passed through the same backbone, except that the latter also leverages the ground truth supervision into the training.
We use the Pearson correlation coefficient-based loss \cite{liu_rppg-mae_2024} to constrain the same rhythm periodicity between the estimated rPPG and the ground truth. 
% The other variants in Table \ref{table4} are trained in the same setup. 
{VL-phys (0\%+100\%) indicates the fully-supervised setting, in which all samples in the training set are leveraged for supervised learning.}
% \ZJ{Compared to the results of supervised methods shown in Table \ref{table1},
% Additionally, we present the performance of the two best-performing supervised methods (\ie, Dual-TL \cite{qian_dual-path_2023} and Dual-Gan \cite{lu_dual-gan_2021}) on UBFC-rPPG and MMVS datasets in Table \ref{table4}, respectively. 
% we can observe that, in this semi-supervised setting, when partial ground truth signals are available, 
First, we can observe that, in the semi-supervised setting, when a partial of ground truth signals is available, VL-phys shows a significant performance improvement, even surpassing the best-performing supervised methods presented in Table \ref{table1}. For instance, with only 30\% of the samples provided with ground truth signals, VL-phys achieves an MAE of 0.13, 0.16, 4.86 and 1.93 across four datasets, respectively, significantly outperforming the best supervised methods. Second, an increase in labeled data from VL-phys to VL-phys (50\%+50\%) yields superior outcomes, suggesting that more ground truth can effectively enhance the learning of the periodicity of rPPG signals. 
% \ZJ{Finally, the performance of Ours (70\%+30\%) is on par with Ours (60\%+40\%), suggesting that performance stabilizes when more than 30\% of the samples are provided with ground truths.} 
{Third, the fully-supervised variant VL-phys (0\%+100\%) achieves the best performance. 
These results demonstrate that our method can be effectively adapted to a semi-supervised or fully-supervised framework and exhibits superior performance in scenarios where some labeled videos are available.}

% Thanks for your comment. In Table IV below, we present the performance of VL-phys under more semi-supervised settings, including VL-phys(60%+40%) and VL-phys(50%+50%). We can see the proposed method indeed reaches the performance saturation point for VL-phys(70%+30%), as further increasing the proportion of labeled data does not yield significant performance improvement.

\begin{table*}[!t]
\caption{{Quantitative performance on subjects with or without makeup.}}
\label{table15}
\begin{center}
\setlength{\tabcolsep}{1.5mm}
\footnotesize
\begin{tabular}{c|ccc|ccc}
\toprule
\multirow{2}*{Method} & \multicolumn{3}{c|}{MMVS-w/ makeup}&\multicolumn{3}{c}{MMVS-w/o makeup}\\ 
&MAE$\downarrow$  &RMSE$\downarrow$  &$r\uparrow$  &MAE$\downarrow$  &RMSE$\downarrow$  &$r\uparrow$\\ 
\midrule
{Gideon} \etal &{3.90} &{5.62} &{0.92}  &{3.39}  &{4.69}  &{0.93}\\
{VL-phys}  &{2.47}  &{3.78}  &{0.94}  &{2.15}  &{3.30}  &{0.95}\\
\bottomrule
\end{tabular}
\end{center}
\end{table*}

\begin{figure}[t]
\begin{center}
\begin{tabular}{cc}
\includegraphics[width=2in]{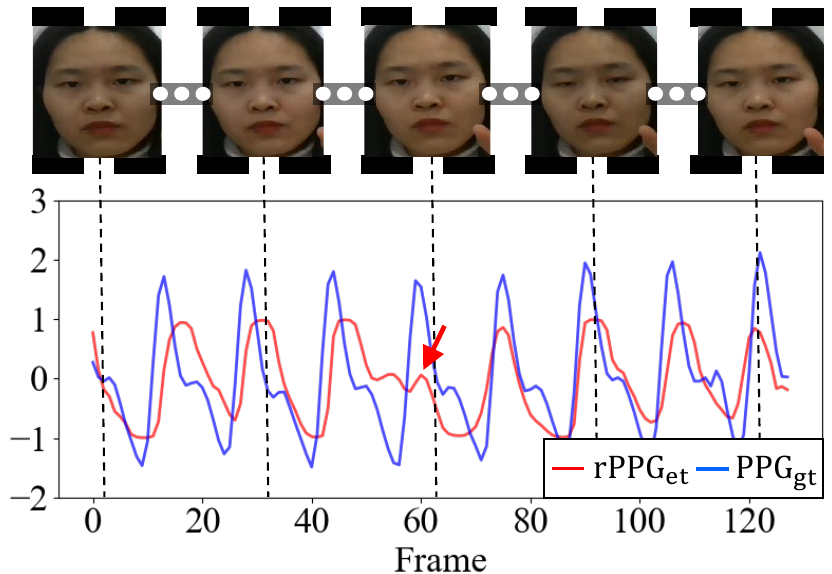} &
\includegraphics[width=2in]{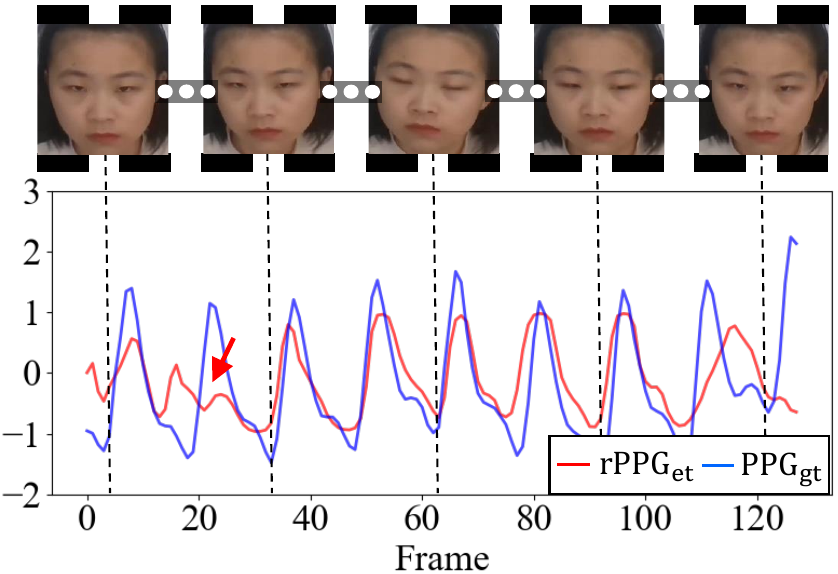} \\
\scriptsize (a) & \scriptsize (b)\\
\end{tabular}
\end{center}
\vspace{-0.1in}
\caption{{Qualitative performance on subjects with makeup.}}
\label{fig14}
\end{figure} 

\subsection{Effect of Makeup}

{We further assess the impact of makeup on our method. Following} \cite{yue_facial_2023}, {we divide the MMVS dataset into two subsets: MMVS-w/ makeup and MMVS-w/o makeup. Each subset comprises a training set and a test set, which are used for model training and evaluation, respectively. The HR estimation results are summarized in Table} \ref{table15}. {We observe that the performance on MMVS-w/o makeup surpasses that on MMVS-w/ makeup: \eg the MAE is 2.15 for the former and 2.47 for the latter. Additionally, compared to} \cite{gideon_way_2021}, {which exhibits a MAE difference of 0.51 between the two subsets, the influence of makeup on our method is smaller (\ie 0.32 MAE difference).} Fig.~\ref{fig14} {shows two estimated rPPG signals and their ground truth signals from subjects with makeup. We can see that some signal peaks are not generated (red arrows). This is because makeup can cover the skin color temporal variations, thereby hindering the extraction of physiological clues for accurate rPPG estimation.}

\section{Ablation study}\label{sec5}

We conduct ablation study for intra-dataset HR estimation on the UBFC-rPPG and MMVS datasets.

\subsection{Text modality}\label{sec6.1}

In our method, we for the first time leverage text prompts to enhance the vision encoder's understanding of frequency-related visual attributes
% frequency patterns in periodic color variations 
for rPPG estimation. We now investigate the effectiveness of leveraging the text modality. In Table \ref{table5}, we denote VL-phys w/o text as a variant where the text modality is removed from the framework. In this way, the STMaps for positive and negative samples are directly fed into the vision encoder for rPPG estimation without constructing the vision-text pairs. The proposed TVR module, vision-text contrastive loss $L_{vtc}$, and reconstruction loss $L_{r}$ are removed due to the absence of the text modality. The results show a significant increase in MAE, rising to 0.57 on the UBFC-rPPG dataset and 2.88 on the MMVS dataset, attesting the usage of the text modality in this task.

\textbf{Text-guided visual reconstruction task.} Text modality is leveraged into the {frequency-related} generative learning task, \ie, text-guided visual reconstruction task, to optimize the vision encoder to reconstruct the C-STMaps.
% to understand the color variation frequency within the vision modality. 
Specifically, a TVR module and a reconstruction loss $L_{r}$ are designed to recover the masked patches of C-STMaps based on the linguistic clues provided by their associated text prompts. If we remove this task from our framework, the MAE and RMSE will be increased for HR estimation (see Table \ref{table5}: VL-phys w/o TVR). This observation suggests that 
% text-guided masked image modeling task 
this task
can effectively encourage the interaction between vision and text modalities, enabling the vision encoder to capture accurate periodic patterns in temporal variations of skin colors.

\textbf{Vision-text contrastive loss.} We further validate the effectiveness of the vision-text contrastive loss $L_{vtc}$ by ablating it in our framework, see Table \ref{table5}. A clear performance drop can be observed by VL-phys w/o $L_{vtc}$. This justifies our primary motivation of performing vision-text contrastive learning to boost the performance.

\begin{table}[!t]
\caption{Ablation study on the text modality.}
\label{table5}
\setlength{\tabcolsep}{1.5mm}
\begin{tabular}{c|ccc|ccc}
\toprule

\multirow{2}*{Method}  &\multicolumn{3}{c|}{UBFC-rPPG}& \multicolumn{3}{c}{MMVS}\\ 

&MAE$\downarrow$  &RMSE$\downarrow$  &$\rho\uparrow$  & MAE$\downarrow$  &RMSE$\downarrow$  &$\rho\uparrow$ \\ 
\midrule

VL-phys &\textbf{0.28} &\textbf{0.60} &\textbf{0.99} &\textbf{2.33} &\textbf{3.76} &\textbf{0.95}\\
VL-phys w/o text &0.57 &0.96 &0.99 &2.88 &4.07 &0.94\\
VL-phys w/o TVR &0.41 &0.72 &0.99 &2.42 &3.83 &0.95\\
VL-phys w/o $L_{vtc}$ &0.50 &0.94 &0.99 &2.70 &3.91 &0.94\\

% Ours w/o $L_{fc}$ &1.78 &2.63 &0.95 &4.75 &6.59 &0.86\\
% Ours w/o $L_{fr}$ &0.51 &0.97 &0.99 &2.66 &3.87 &0.94\\
% $L_{fr}\rightarrow L_{frc}$ &0.34 &0.61 &0.99 &2.43 &3.82 &0.95\\
% $L_{fr}\rightarrow L_{fr-pw}$ &0.37 &0.64 &0.99 &2.40 &3.81 &0.95\\
\bottomrule
\end{tabular}
\end{table}

\begin{table}[!t]
\caption{Ablation study on the vision and text encoders.}
\vspace{-0.1in}
\label{table6}
\setlength{\tabcolsep}{0.9mm}
\begin{tabular}{c|ccc|ccc}
\toprule
\multirow{2}*{Method}  &\multicolumn{3}{c|}{UBFC-rPPG}& \multicolumn{3}{c}{MMVS}\\ 

&MAE$\downarrow$  &RMSE$\downarrow$  &$\rho\uparrow$  & MAE$\downarrow$  &RMSE$\downarrow$  &$\rho\uparrow$ \\ 
\midrule
VL-phys &0.28 &\textbf{0.60} &\textbf{0.99} &2.33 &3.76 &\textbf{0.95}\\
VL-phys(CLIP) &0.49 &0.79 &0.99 &2.83 &4.07 &0.94\\
VL-phys(COCA)  &0.47 &0.74 &0.99 &2.65 &3.90 &0.95\\
VL-phys(ALBEF)&0.48 &0.80 &0.99 &2.69 &4.01 &0.95\\
VL-phys(BLIP) &0.41 &0.71 &0.99 &2.45 &3.81 &0.95\\
VL-phys(BLIP2)&0.41 &0.70 &0.99 &2.47 &3.82 &0.95\\
% Ours(MiniGPT4) \cite{zhu_minigpt-4_nodate}&0.44 &0.78 &0.99 &2.44 &3.98 &0.95\\
% Ours(MiniGPTv2) \cite{chen_minigpt-v2_nodate} &0.39 &0.71 &0.99 &2.39 &3.83 &0.95\\
% VL-phys(LLAVA) \cite{liu_visual_2024} &0.45 &0.76 &0.99 &2.43 &3.88 &0.95\\
VL-phys(All-in-one) &\textbf{0.25} &0.61 &0.99 &2.36 &3.79 &0.95\\
VL-phys(CLIP-ViP)  &0.26 &0.65 &0.99 &2.34 &3.78 &0.95\\
VL-phys w/o pw &0.55 &0.83 &0.99 &2.90 &4.19 &0.94\\
% {VL-phys(ViT-B+BERT)}	&{0.53}	&{0.83}	&{0.99}	&{2.84}	&{4.10}	&{0.94}\\
% {VL-phys(ViT-B+RoBERTa)}	&{0.50}	&{0.82}	&{0.99}	&{2.79}	&{3.98}	&{0.95}\\
% {VL-phys(ViT-L+BERT)}	&{0.47}	&{0.82}	&{0.99}	&{2.79}	&{3.95}	&{0.95}\\
% {VL-phys(ViT-L+RoBERTa)}	&{0.40}	&{0.79}	&{0.99}	&{2.75}	&{3.88}	&{0.95}\\
VL-phys-MLP &1.57 &2.91 &0.97 &4.03 &6.17 &0.91\\
VL-phys-MLM &0.32 &0.66 &0.99 &2.38 &3.90 &0.95\\
VL-phys(BLIP)-MLM &0.50 &0.73 &0.99 &2.44 &3.87 &0.95\\
VL-phys-MLM(r) &0.27 &0.60 &0.99 &\textbf{2.31} &\textbf{3.75} &0.95\\
{VL-phys(VPT)}	&{0.61}	&{0.95}	&{0.99}	&{2.96}	&{4.38}	&{0.94}\\
{VL-phys(CoOp)}	&{0.59}	&{0.87}	&{0.99}	&{2.91}	&{4.11}	&{0.94}\\
{VL-phys(Adapter)}	&{0.54}	&{0.86}	&{0.99}	&{2.84}	&{4.10}	&{0.94}\\
% Ours(BLIP)-MLM-r &0.39 &0.68 &0.99 &2.47 &3.84 &0.95\\

\bottomrule
\end{tabular}
\end{table}

\subsection{Vision and text encoders}\label{sec6.2}

We employ the pre-trained vision and text encoders from VindLU \cite{cheng_vindlu_2023} to extract features from the generated vision-text pairs. Table \ref{table6} details the performance of variants using encoders from other widely-used VLMs, \ie, CLIP \cite{radford_learning_2021}, COCA \cite{yu_coca_nodate}, ALBEF \cite{li_align_2021}, BLIP \cite{li_blip_2022}, BLIP2 \cite{li_blip-2_2023}, All-in-one \cite{wang_all_2023}, CLIP-ViP \cite{xue_clip-vip_2023}. For example, VL-phys(CLIP) indicates a variant in which we use the pre-trained vision and text encoders from CLIP.

First, it is evident that all of these variants outperform the state of the art self-supervised rPPG estimation methods in Table \ref{table1}. This essentially validates that the VLMs fine-tuned with our VL-phy possess more powerful ability to capture the periodic variation of skin color for rPPG estimation.
% incorporating text modality can enhance the vision encoder's understanding of periodic color variation patterns.

Second, one can see that our default setting with VindLU exhibits comparable (slightly superior) performance with VL-phys(All-in-one) and VL-phys(CLIP-ViP). While other variants perform inferior. We attribute this to the comprehensive training of VindLU, All-in-one, and CLIP-ViP across large-scale video-text datasets. In contrast, other VLMs such as BLIP and COCA are pre-trained on image-text datasets, which are not optimum for our video-based task. The extensive coverage of dynamic scenes and contexts allows VindLU, All-in-one, and CLIP-ViP to develop richer spatio-temporal feature encoding capabilities.
%Consequently, fine-tuning their pre-trained encoders to capture periodic temporal patterns of color variations across frames becomes relatively straightforward.

\textbf{Pre-trained model weights.} We present another variant that does not use the pre-trained weights from VindLU, but instead to initialize the weights of the encoders randomly. This variant is denoted as VL-phys w/o pw in Table \ref{table6}. The results get worse to 0.55 and 2.39 in MAE on two datasets, respectively. This indicates that pre-training is important for the understanding of the text prompts describing variational visual details.
% fine-tuning encoders with a basic temporal modeling capability can benefit rPPG estimation.}

% {\textbf{Independently pre-trained encoders.} We present experimental results of VL-phys using independently pre-trained vision encoders (\ie ViT-B and ViT-L) and text encoders (\ie BERT and RoBERTa) to extract embeddings from the generated frequency-oriented vision-text pairs. As shown in Table} \ref{table6}, {we can see these variants (\ie VL-phys(ViT-B+BERT), VL-phys(ViT-B+RoBERTa), VL-phys(ViT-L+BERT), VL-phys(ViT-L+RoBERTa)) perform inferior to the original VL-phys, which utilizes encoders derived from pre-trained VLMs. The reason is that the pre-trained VLMs are optimized to align vision and language modalities effectively through large-scale multi-modal pretraining. Their encoders are inherently capable of understanding basic concepts embedded in our text prompts, such as “horizontal”, “left side”, and “right side”, and can effectively capture the relationships between these prompts and the visual features of C-STMap. In contrast, independently pre-trained vision and text encoders lack such cross-modal alignment, making it more challenging to understand the text prompts describing variational visual details. Therefore, we employ the vision and text encoders from pre-trained VLMs.}

\textbf{Textual embedding extraction.} We verify the benefits of using text encoder BERT in our framework. Given that BERT has approximately 110 million parameters, we introduce an alternative light-weight variant where the text encoder is only an MLP with five linear layers. This encoder exclusively processes the [ratio] token from the text, and its output is directly employed for vision-text contrastive learning and text-guided visual reconstruction. This variant is denoted as VL-phys-MLP in Table \ref{table6}. We can observe severe performance degradation by this variant, attesting the importance of using an established text encoder. Moreover, compared to using the single token of [ratio], our designed text prompt provides richer semantics describing the skin color temporal variation. %This enhances the fine-grained alignment of vision-text pairs, hence performs better.

\textbf{Image-guided masked language modeling task.} Image-guided masked language modeling task is another kind of generative learning task that could potentially be integrated into our VL-phys. Given a vision-text pair, it aims to predict several randomly masked tokens in the text prompt conditioned on visual tokens and the remaining unmasked text tokens. This task differs from our designed text-guided visual reconstruction task, which focuses on recovering visual patches. We have incorporated this auxiliary task into our framework to evaluate it. This variant is denoted as VL-phys-MLM in Table \ref{table6}. It should be noted that popular VLMs typically leverage a multimodal fusion encoder to fuse multimodal cues and reconstruct the masked tokens based on vision-grounded textual embeddings. We employ the pre-trained multimodal fusion encoder from VindLU for this purpose. We can observe a degraded performance for this variant.
% , with an accuracy of 82.15\% in reconstructing correct tokens. 
Similar findings are also noted from VL-phys(BLIP) to VL-phys(BLIP)-MLM.
% , with an accuracy of 80.73\%. 
The issue primarily arises due to the uninformative tokens within the text prompts, such as "the", "of", "is" and "as". When these tokens are masked instead of more informative ones like the [ratio] token, the model tends to predict them using solely linguistic cues, rather than trying to find the answer from the visual modality. 
% Our finding is consistent with that reported in \cite{Gou_2023_CVPR}.

Furthermore, we have improved VL-phys-MLM by setting the masked tokens in the text prompts consistently to the most important [ratio] token, denoted as VL-phys-MLM(r) in Table \ref{table6}. This variant outperforms VL-phys-MLM and indeed shows some negligible improvement to VL-phys. However, considering the additional computation incurred by the multimodal fusion encoder, we recommend not to use this variant by default.

{\textbf{Fine-tuning methods.} During the training stage of VL-phys, we fine-tune the pre-trained vision and text encoders of VLM using the full fine-tuning strategy. We present three variants that do not use the full fine-tuning strategy, but instead use parameter-efficient fine-tuning techniques for network optimization, including prompt-tuning} (\ie {VPT} \cite{jia2022visual} {and CoOp} \cite{zhou2022learning}) {and adapter-tuning} \cite{chen2022adaptformer}. {These variants are denoted as VL-phys (VPT), VL-phys (CoOp), and VL-phys (Adapter) in Table} \ref{table6}. {We can see they achieve inferior performance compared to the original VL-phys. This can be attributed to the difficulty of fine-tuning the encoders to effectively understand frequency-related attributes. Unlike parameter-efficient fine-tuning, full fine-tuning enables the encoders to comprehensively digest the frequency-related knowledge of skin color temporal variation.}

%(approximately an extra three hours of training time on the MMVS dataset)

% with the masked token consistently set as the [ratio] token in the texts. Considering the computation increase introduced by the multimodal fusion encoder, we do not use this task.}

% Additionally, BLIP may have developed an understanding of the concept of frequency through the training process with large amounts of vision-text pairs, which is crucial for capturing temporal patterns and their periodic variations within images. Consequently, it shows better estimation performance. Consequently, it shows better estimation performance., which encompass a variety of scenarios ALBEF is trained on only four of them, while COCA is pre-trained on just one

% the deep text encoder BERT pre-trained on extensive vision-text pairs.}
% 

\begin{table}[!t]
\caption{Ablation study on loss functions.}
\label{table7}
\setlength{\tabcolsep}{1.5mm}
\begin{tabular}{c|ccc|ccc}
\toprule

\multirow{2}*{Method}  &\multicolumn{3}{c|}{UBFC-rPPG}& \multicolumn{3}{c}{MMVS}\\ 

&MAE$\downarrow$  &RMSE$\downarrow$  &$\rho\uparrow$  & MAE$\downarrow$  &RMSE$\downarrow$  &$\rho\uparrow$ \\ 
\midrule

VL-phys &\textbf{0.28} &\textbf{0.60} &\textbf{0.99} &\textbf{2.33} &\textbf{3.76} &\textbf{0.95}\\
VL-phys w/o $L_{fc}$ &1.78 &2.63 &0.95 &4.75 &6.59 &0.86\\
VL-phys w/o $L_{fr}$ &0.51 &0.97 &0.99 &2.66 &3.87 &0.94\\
$L_{fr}\rightarrow L_{afr}$ &0.33 &0.60 &0.99 &2.42 &3.81 &0.95\\
% $L_{fr}\rightarrow L_{fr-pw}$ &0.36 &0.63 &0.99 &2.39 &3.80 &0.95\\
\bottomrule
\end{tabular}
\end{table}

\subsection{Losses}\label{sec6.3}

\textbf{Frequency contrastive loss.} Frequency contrastive loss $L_{fc}$ is the most important loss in existing self-supervised remote physiological measurement methods \cite{yang_simper_2023,yue_facial_2023,gideon_way_2021,sun_contrast-phys_2024}. It enforces the signal frequency similarities among positive samples and dissimilarities between positive and negative samples in the feature space. In Table \ref{table7} we use VL-phys w/o $L_{fc}$ to denote a variant without using $L_{fc}$ in our framework. We can see that the MAE and RMSE are substantially increased while $\rho$ is substantially decreased from the original framework, \eg, +2.42, +2.83 and -0.09 on MMVS, which proves the benefits of adding $L_{fc}$.

\textbf{Frequency ranking loss.} We verify the effectiveness of the proposed frequency ranking loss $L_{fr}$ (Sec.~\ref{sec3.6}). First, as shown in Table \ref{table7}, omitting $L_{fr}$ leads to deteriorated results on both datasets.
% Without it, the vision encoder struggles to determine the relative frequency ratios of color variations among different video samples, leading to frequency estimation errors in the predicted rPPG signals.
% Next, we explore a variant of $L_{fr}$. 1) The frequency ratio consistency loss $L_{frc}$ \cite{yue_facial_2023} appears to be a viable alternative to $L_{fr}$. It uses an absolute term to constrain the signal frequency ratios between positive and negative video samples to be consistent. However, unlike $L_{frc}$, we design our $L_{fr}$ as a weaker constraint to avoid overfitting. If we replace $L_{fr}$ with $L_{frc}$ in our framework (denoted by $L_{fr}\rightarrow L_{frc}$ in Table \ref{table7}), 
Next, $L_{fr}$ optimizes the relative frequency ranks of predicted positive and negative signals rather than their absolute frequency differences to avoid overfitting. If we instead constrains the signal frequency ratios between these signals to be consistent with the absolute ground truth ratios $R$, denoted by $L_{fr}\rightarrow L_{afr}$ in Table \ref{table7}, the performance gets worse, \eg, +0.05 and +0.09 in MAE on two datasets, respectively.
% 2) We design $L_{fr}$ inspired by the pairwise ranking loss introduced in \cite{chen_ranking_2009}. \cite{chen_ranking_2009} also proposed a pointwise ranking loss, which can be used to optimize the predicted frequency ranks of rPPG signals to align with the assigned ground truth ranks. If we instead optimize our method using it, denoted by $L_{fr}\rightarrow L_{fr-pw}$ in Table \ref{table7}, the performance deteriorates compared to the original $L_{fr}$: \eg, the MAE and RMSE increase by +0.08 and +0.03 on UBFC-rPPG. These observations confirm the superiority of $L_{fr}$.

\begin{table*}[!t]
\caption{Ablation study on different text prompts.}
\vspace{-0.1in}
\label{table8}
\begin{center}
\setlength{\tabcolsep}{1mm}
\hspace*{-2cm} % 负值表示向左移动，调整数值以达到合适的位置
\footnotesize % 调整表格的字号为小一号
\begin{tabular}{c|c|ccc|ccc}
\toprule
\multirow{2}*{Method} &\multirow{2}*{Text template} &\multicolumn{3}{c|}{UBFC-rPPG}& \multicolumn{3}{c}{MMVS}\\ 

& &MAE$\downarrow$  &RMSE$\downarrow$  &$\rho\uparrow$  & MAE$\downarrow$  &RMSE$\downarrow$  &$\rho\uparrow$ \\ 
\midrule
VL-phys & \parbox{9cm}{\centering The frequency of the horizontal color variation on the left/right side is [ratio] times of that on the right/left side of the image.} &0.28 &\textbf{0.60} &\textbf{0.99} &\textbf{2.33} &3.76 &\textbf{0.95}\\
\midrule
VL-phys-T1 &\parbox{9cm}{\centering The frequency of the horizontal color variation on the left/right side is [ratio] as high as on the right/left side of the image.} &0.33 &0.65 &0.99 &2.35 &\textbf{3.71} &0.95\\
\midrule
VL-phys-T2 &\parbox{9cm}{\centering The frequency of horizontal color variation on the left/right side is [ratio] as great as on the right/left side of the image.} &\textbf{0.27} &0.63 &0.99 &2.39 &3.79 &0.95\\

\midrule
% Ours-T3 &\parbox{9cm}{\centering The horizontal color changes on the left/right side is [ratio] as fast as on the right/left side of the image.} &0.55 &0.92 &0.99 &2.82 &4.03 &0.94\\
VL-phys-T3 &\parbox{9cm}{\centering The horizontal color variation on the left/right side is [ratio] times as fast as on the right/left side of the image.} &0.31 &0.64 &0.99 &2.33 &3.80 &0.95\\
\midrule
% Ours-An &default &0.41 &0.73 &0.99 &2.48 &3.90 &0.94\\
VL-phys-Video &\parbox{9cm}{\centering The frequency of skin color temporal variation in the first/second half of the video is [ratio] of that in the second/first half.} &0.57 &0.96 &0.99 &3.01 &4.22 &0.93\\
\bottomrule
\end{tabular}
\end{center}
\end{table*}

\subsection{Text prompt design}\label{sec6.4}

We analyze the influence of text prompt design in Table \ref{table8}. We denote VL-phys-T1, VL-phys-T2, VL-phys-T3 as three variants,
% training with other three alternative templates, respectively. 
each uses an alternative prompt.
% generated from GPT-4.
We observe that these varints show comparable performance to the original VL-phys, indicating that our approach remains effective as long as the text prompt is grammatically correct and can accurately depict the relative frequency ratios of skin color temporal variations in C-STMaps. This robustness highlights our method's ability to adapt to various text prompts while consistently capturing frequency-related attributes for rPPG estimation.

% Ours-T3 has the worst performance across all metrics. We attribute this to the pre-trained text encoder of VindLU, which struggles to interpret semantics related to the "fast" or "slow" speed of events, as such descriptions are rare in its pre-training datasets.

Additionally, we introduce VL-phys-Video as another variant without generating spatio-temporal maps (STMaps) from positive and negative video samples. In this variant, we concatenate the positive and negative video along the temporal dimension to obtain a cross-sample video. We then define its corresponding text prompt using template shown in Table \ref{table8}. We can see the MAE respectively increases to 0.57 and 3.01 on two datasets by this variant. This can be owed to the potential noises and distractions when directly using the original videos. 
%some pixels may belong to the background that do not contribute to rPPG estimation.
% ; while some other pixels, despite on the face, are insensitive to blood pulsation (\eg, eyes and mouth). 
%The noises contained in these pixels would adversely affect the vision encoder to extract pulsation information, resulting in poor quality of the predicted rPPG signals. 
In contrast, STMaps retain only pixels from detected facial ROIs while masking other irrelevant pixels in the
videos. They can effectively highlight the temporal variation of skin color.

\begin{table}[!t]
\caption{Zero-shot HR estimation performance of pre-trained VLMs.}
\vspace{-0.1in}
\label{table12}
\setlength{\tabcolsep}{1.5mm}
\begin{tabular}{c|ccc|ccc}
\toprule
\multirow{2}*{Method}  &\multicolumn{3}{c|}{UBFC-rPPG}& \multicolumn{3}{c}{MMVS}\\ 

&MAE$\downarrow$  &RMSE$\downarrow$  &$\rho\uparrow$  & MAE$\downarrow$  &RMSE$\downarrow$  &$\rho\uparrow$ \\ 
\midrule

VL-phys &\textbf{0.28} &\textbf{0.60} &\textbf{0.99} &\textbf{2.33} &\textbf{3.76} &\textbf{0.95}\\
CLIP &9.82 &11.37 &0.15 &10.40 &13.55 &0.18\\
BLIP &8.46 &9.90 &0.28 &10.37 &14.19 &0.20\\
ALBEF &8.71 &10.22 &0.23 &11.65 &14.21 &0.15\\
\bottomrule
\end{tabular}
\end{table}

\subsection{Zero-shot rPPG estimation using pretrained VLMs}\label{sec6.5}

We investigate the zero-shot rPPG estimation performance using off-the-shelf pre-trained VLMs (CLIP, BLIP, ALBEF) corresponding to the discussion in Sec.~\ref{sec1}. Initially, we define a frequency bin [0.5, 1.0, 1.5, 2.0, 2.5] and create text prompts to describe the frequency of skin color temporal variation. The text template is defined as "the frequency of skin color temporal variation is [ratio] hertz in the video." We respectively replace the [ratio] token with each frequency value in the bin and then calculate the similarity between the embeddings of the text prompt and the given video. After calculating the similarity score for each frequency value, we linearly combine them to obtain the final prediction value. 
% We then convert the results to HR for performance comparison with our VL-phys. 
We then multiply the results by 60\footnote{The frequency value should be multiplied by 60 to convert it from hertz to HR in beats per minute (bpm).} to calculate the HR for performance comparison with our VL-phys. As shown in Table \ref{table12}, we can see a substantial performance gap between using VL-phys and off-the-shelf pre-trained VLMs, highlighting their difficulties in capturing periodic visual patterns. These findings also reveal that VLMs need to be re-trained or fine-tuned based on specific vision-text pairs, otherwise they would struggle to understand the attributes related to frequency.
% the effectiveness of our method.}

\subsection{Visual embeddings for rPPG estimation }\label{sec6.6}
We adopt the embeddings of the [CLS] tokens from STMaps to predict rPPG signals. We can also aggregate information from local patches of STMaps for the same rPPG estimation purpose. In Table \ref{table11}, we denote VL-phys(g$\rightarrow$l) as a variant where the rPPG signals are predicted from the average of patch embeddings from STMaps. VL-phys(g+l) is another variant that concatenates the global [CLS] token embedding and the average of patch embeddings for rPPG estimation. VL-phys(g+l) show similar performance with the original VL-phys. Considering the computation increase, we opt to use the [CLS] tokens only.

\begin{table}[!t]
\caption{Ablation study on visual embeddings for rPPG estimation.}
\vspace{-0.1in}
\label{table11}
\setlength{\tabcolsep}{1.5mm}
\begin{tabular}{c|ccc|ccc}
\toprule
\multirow{2}*{Method}  &\multicolumn{3}{c|}{UBFC-rPPG}& \multicolumn{3}{c}{MMVS}\\ 

&MAE$\downarrow$  &RMSE$\downarrow$  &$\rho\uparrow$  & MAE$\downarrow$  &RMSE$\downarrow$  &$\rho\uparrow$ \\ 
\midrule
VL-phys &0.28 &\textbf{0.60} &\textbf{0.99} &\textbf{2.33} &\textbf{3.76} &\textbf{0.95}\\
VL-phys(g$\rightarrow$l) &0.31 &0.62 &0.99 &2.40 &3.81 &0.95\\
VL-phys(g+l) &\textbf{0.27} &0.60 &0.99 &2.35 &3.77 &0.95\\
\bottomrule
\end{tabular}
\end{table}

\begin{table}[!t]
\caption{{Ablation study on the negative sample generation method.}}
\vspace{-0.1in}
\label{table16}
\setlength{\tabcolsep}{1.5mm}
\begin{tabular}{c|ccc|ccc}
\toprule
\multirow{2}*{Method}  &\multicolumn{3}{c|}{UBFC-rPPG}& \multicolumn{3}{c}{MMVS}\\ 
&MAE$\downarrow$  &RMSE$\downarrow$  &$\rho\uparrow$  & MAE$\downarrow$  &RMSE$\downarrow$  &$\rho\uparrow$ \\ 
\midrule
{VL-phys} &{\textbf{0.28}} &{\textbf{0.60}} &{\textbf{0.99}} &{\textbf{2.33}} &{\textbf{3.76}} &{\textbf{0.95}}\\
{LFA$\rightarrow$NST} &{0.29} &{0.62} &{0.99} &{2.37} &{3.79} &{0.95}\\
{VL-phys(br)}	 &{0.79} &{1.34}	 &{0.98}	 &{3.65}	 &{4.81}	 &{0.93}\\
\bottomrule
\end{tabular}
\end{table}

 \begin{figure}[t]
\begin{center}
\begin{tabular}{cc}
\includegraphics[width=1.65in]{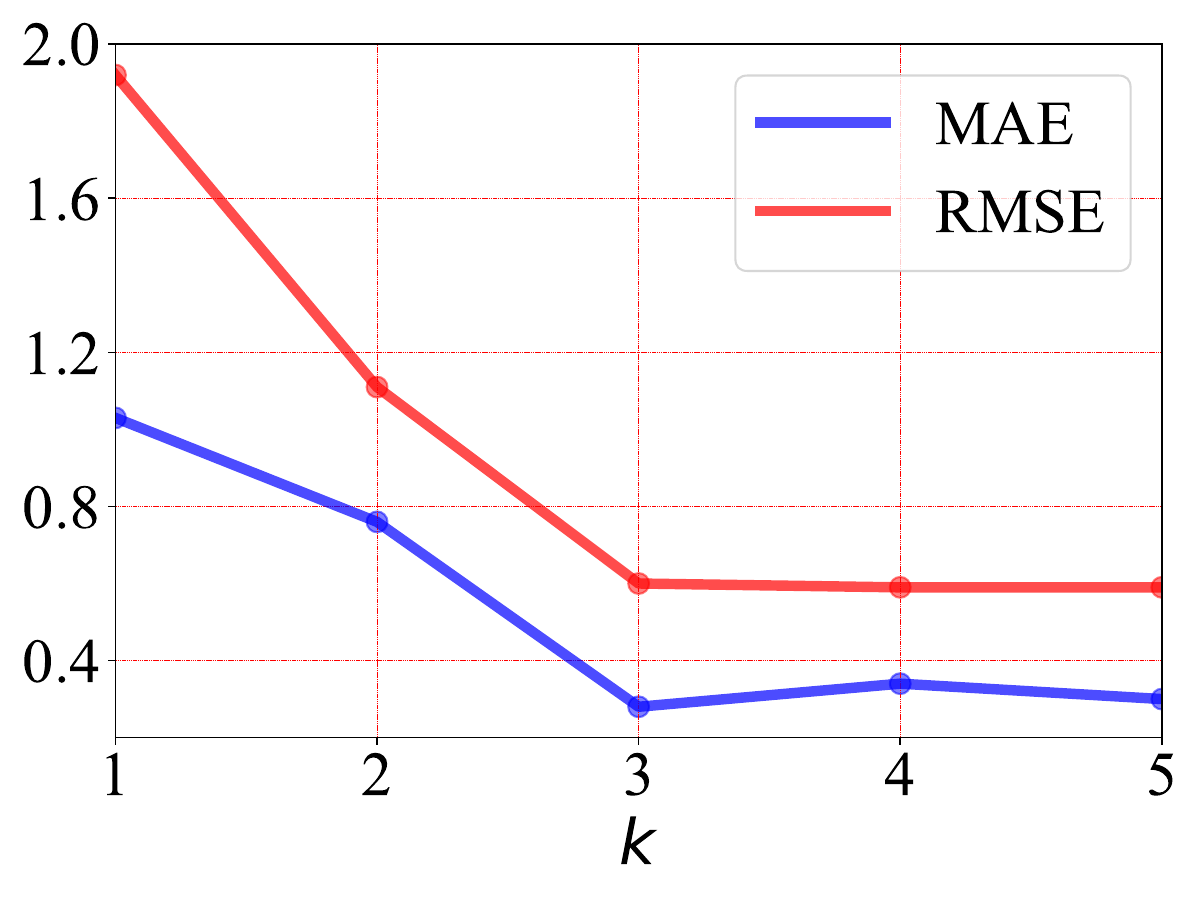} &
\includegraphics[width=1.65in]{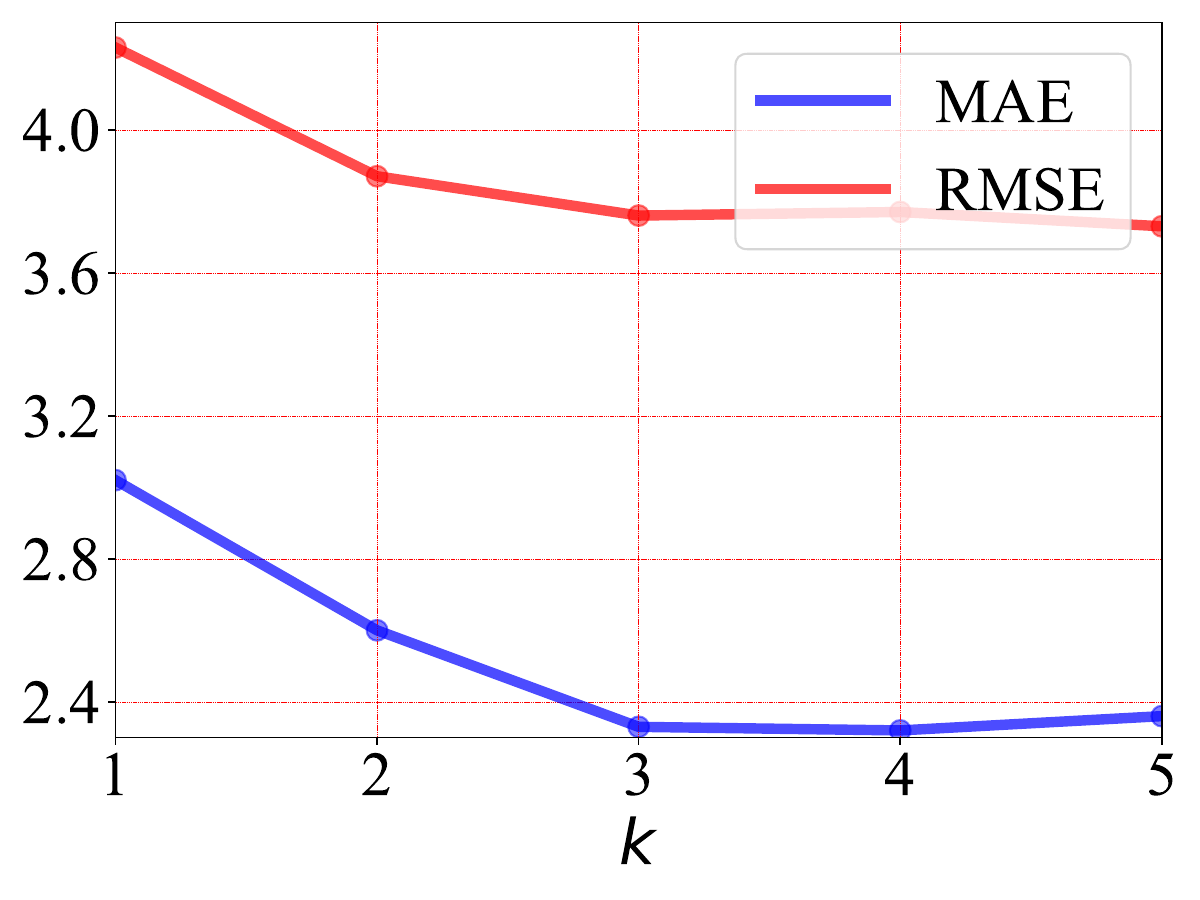} \\
(a) UBFC & (b) MMVS\\
\end{tabular}
\end{center}
\caption{Parameter variation on the number of negative samples.}

\label{fig8}
\end{figure}

% \begin{table}[!t]
% \caption{Ablation study on different number of negative samples.}
% \vspace{-0.1in}
% \label{table9}
% \begin{center}
% \setlength{\tabcolsep}{1.5mm}
% \begin{tabular}{c|ccc|ccc}
% \toprule
% \multirow{2}*{Method}  &\multicolumn{3}{c|}{UBFC-rPPG}& \multicolumn{3}{c}{MMVS}\\ 

% &MAE$\downarrow$  &RMSE$\downarrow$  &$r\uparrow$  & MAE$\downarrow$  &RMSE$\downarrow$  &$r\uparrow$ \\ 
% \midrule
% VL-phys ($k = 1$)  &1.03 &1.92 &0.99 &3.02 &4.23 &0.94\\
% VL-phys ($k = 2$)  &0.76 &1.11 &0.99 &2.60 &3.87 &0.94\\
% VL-phys ($k = 3$) &\textbf{0.28} &0.60 &\textbf{0.99} &2.33 &3.76 &\textbf{0.95}\\
% VL-phys ($k = 4$) &0.34 &0.59 &0.99 &\textbf{2.32} &3.77 &0.95\\
% VL-phys ($k = 5$) &0.30 &\textbf{0.59} &0.99 &2.36 &\textbf{3.73} &0.95\\

% \bottomrule
% \end{tabular}
% \end{center}
% \end{table}

\subsection{Generation of negative samples }\label{sec6.7}
{In the framework of VL-phys, we use the LFA module} \cite{yue_facial_2023} {to generate the negative samples. We further verify the model performance using an alternative negative sampling method} \cite{wang_self-supervised_2022}, {where negative samples are generated by altering video speed according to the Nyquist sampling theorem. The results of this variant (denoted as LFA$\rightarrow$NST) are presented in Table} \ref{table16}. {We can see this variant achieves comparable performance with the original VL-phys. This demonstrates the robustness and flexibility of the proposed framework in adapting to different negative sample generation methods.}

{Moreover, we evaluate the effectiveness of our predefined frequency ratio bin used in the LFA module for negative samples generation. According to} \cite{yu2021facial}, {the frequency of skin color temporal variation in real facial videos typically lies within the range of [0.75, 4] Hz. Based on this observation, we predefine a frequency ratio bin with a narrow value range to avoid significant frequency changes between the generated positive and negative samples. This design enables the subsequent vision encoder to focus on distinguishing the samples based on their subtle frequency differences, which is essential for accurately extracting underlying rPPG signals. If the frequency ratio $r$ is instead randomly sampled from a broader range (0, 10), as shown in the variant VL-phys (br) in Table} \ref{table16}, {the MAE and RMSE increase significantly, \eg +1.32 and +1.05 on the MMVS dataset. These results underscore the effectiveness and superiority of our predefined ratio bin.}

We further vary the number of negative video samples in this session. In Fig.~\ref{fig8}, we vary this number $k$ from 1 to 6 and observe that the performance appears to be stable when $k\ge3$. Considering that a big $k$ would increase the computation cost, we select $k=3$ as our default setting.

\begin{figure}[t]
\begin{center}
\begin{tabular}{cc}
\includegraphics[width=1.65in]{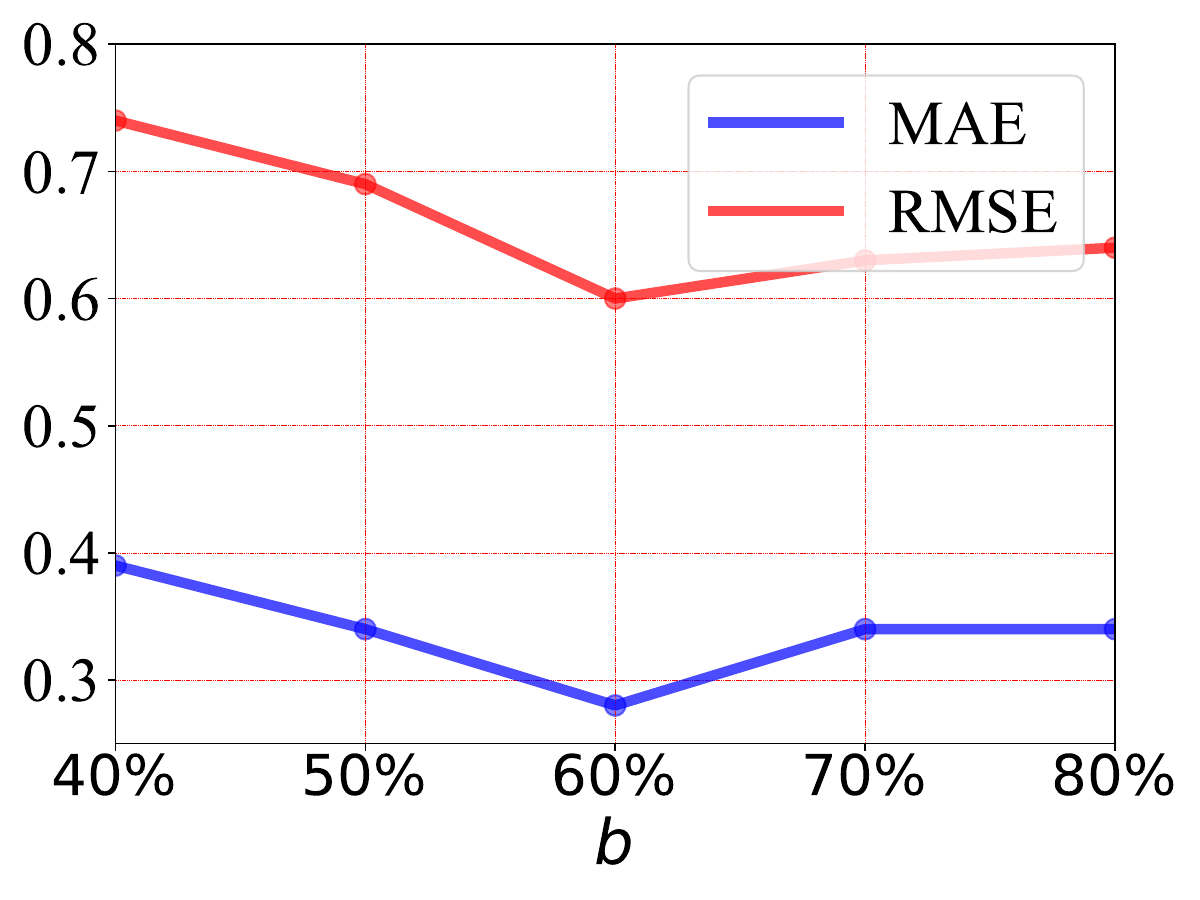} &
\includegraphics[width=1.65in]{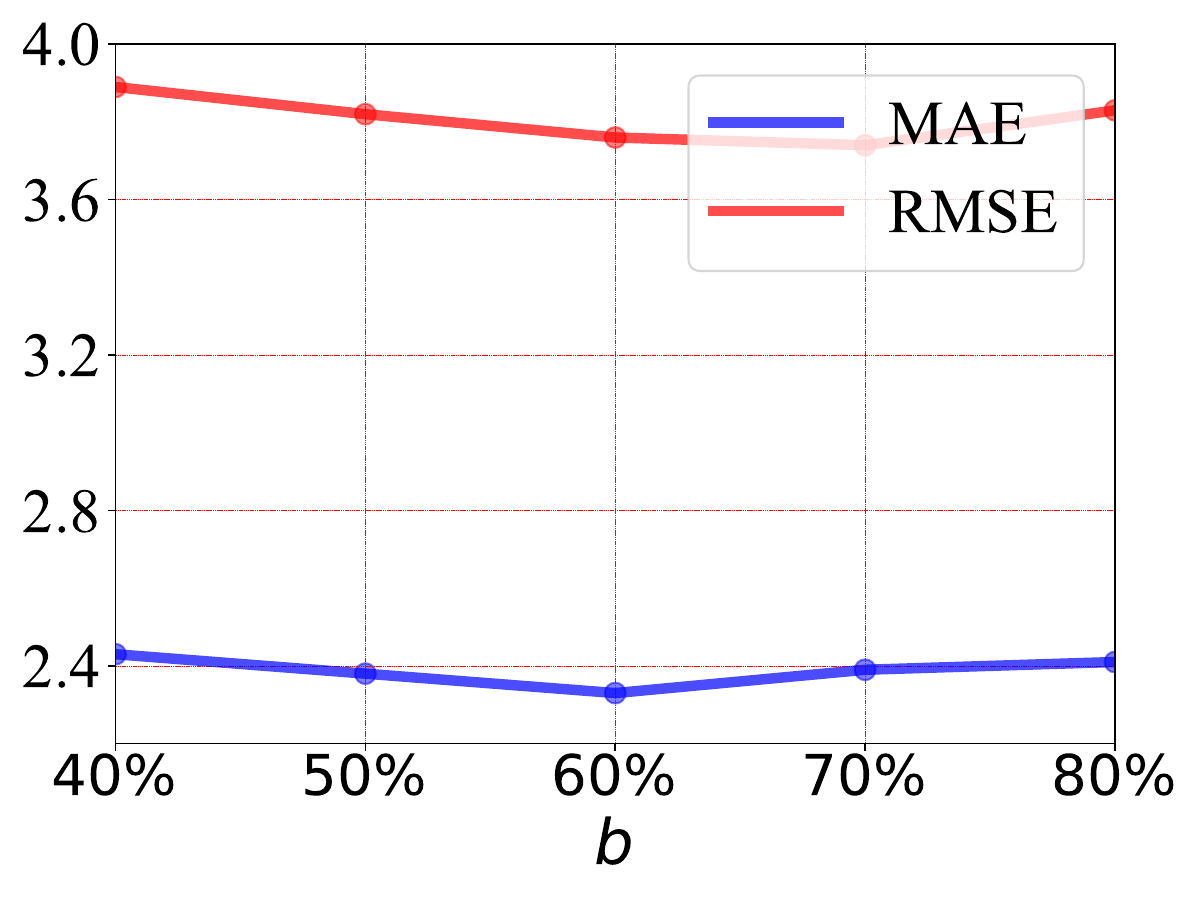} \\
(a) UBFC & (b) MMVS\\
\end{tabular}
\end{center}
\caption{Parameter variation on the masking ratio in C-STMaps.}

\label{fig9}
\end{figure}

% \begin{table}[!t]
% \caption{Ablation study on different masking ratios.}
% \vspace{-0.1in}
% \label{table10}
% \begin{center}
% \setlength{\tabcolsep}{1.5mm}
% \begin{tabular}{c|ccc|ccc}
% \toprule
% \multirow{2}*{Method}  &\multicolumn{3}{c|}{UBFC-rPPG}& \multicolumn{3}{c}{MMVS}\\ 

% &MAE$\downarrow$  &RMSE$\downarrow$  &$r\uparrow$  & MAE$\downarrow$  &RMSE$\downarrow$  &$r\uparrow$ \\ 
% \midrule
% VL-phys ($b = 40\%$)  &0.39 &0.73 &0.99 &2.43 &3.89 &0.95\\
% VL-phys ($b = 50\%$)  &0.34 &0.69 &0.99 &2.38 &3.82 &0.95\\
% VL-phys ($b = 60\%$) &\textbf{0.28} &\textbf{0.60} &\textbf{0.99} &2.33 &3.76 &\textbf{0.95}\\
% VL-phys ($b = 70\%$) &0.34 &0.63 &0.99 &2.39 &\textbf{3.74} &0.95\\
% VL-phys ($b = 80\%$) &0.30 &0.64 &0.99 &\textbf{2.32} &3.83 &0.95\\

% \bottomrule
% \end{tabular}
% \end{center}
% \end{table}

\begin{table}[!t]
\caption{{Ablation study on the spatio-temporal map generation methods.}}
\vspace{-0.1in}
\label{table17}
\setlength{\tabcolsep}{1.5mm}
\begin{tabular}{c|ccc|ccc}
\toprule
\multirow{2}*{Method}  &\multicolumn{3}{c|}{UBFC-rPPG}& \multicolumn{3}{c}{MMVS}\\ 
&MAE$\downarrow$  &RMSE$\downarrow$  &$\rho\uparrow$  & MAE$\downarrow$  &RMSE$\downarrow$  &$\rho\uparrow$ \\ 
\midrule
{VL-phys} &{\textbf{0.28}} &{\textbf{0.60}} &{\textbf{0.99}} &{\textbf{2.33}} &{\textbf{3.76}} &{\textbf{0.95}}\\
{VL-phys(POS)}	&{0.30}	&{0.63}	&{0.99}	&{2.32}	&{3.76}	&{0.95}\\
{VL-phys(CHROM)}	&{0.31}	&{0.63}	&{0.99}	&{2.37}	&{3.78}	&{0.95}\\
\bottomrule
\end{tabular}
\end{table}

\subsection{Generation of STMaps}\label{sec6.8}
{In our framework, we use the spatio-temporal map generation method proposed in} \cite{niu_rhythmnet_2020} and \cite{lu_dual-gan_2021} {to generate the STMaps for positive and negative samples. However, different kinds of STMaps can be generated by using alternative methods discussed in} \cite{liu_rppg-mae_2024}. Table \ref{table17} {presents two variants, VL-phys (POS) and VL-phys (CHROM), where our generated STMaps are replaced by POS-STMaps and CHROM-STMaps, respectively. POS-STMap consists of U and V channels derived from the original facial videos, along with a POS channel generated from the POS algorithm} \cite{wang_algorithmic_2017}. {Similarly, CHROM-STMap also incorporates the U and V channels, but the POS channel is replaced by the CHROM channel generated from the CHROM algorithm} \cite{de_haan_robust_2013}. {We can see these variants have similar performance with the original VL-phys. This indicates that the proposed framework is robust and generalizable across different spatio-temporal map generation methods.}

\subsection{Masking ratio in C-STMaps}\label{sec6.9}
We follow \cite{kwon_masked_2023} to mask non-overlapping patches from contrastive spatio-temporal maps (C-STMaps) based on a masking ratio of $b=60\%$. To explore the impact of this ratio, we also experiment with masking ratios of 40\%, 50\%, 70\%, and 80\%, and report the  performance in Fig.~\ref{fig9}. Our results reveal that lower masking ratios yield inferior results. Maintaining a low masking ratio causes the TVR module to primarily reconstruct masked patches based on cues from the unmasked patches, rather than relying on text prompts. On the other hand, excessively relying on textual embeddings is also not recommended. 
% This is detrimental to achieving fine-grained alignment between vision and text modalities. 
Therefore, we set the $b=60\%$ as default.

\begin{figure}[t]
\begin{center}
\begin{tabular}{cc}
\includegraphics[width=1.65in]{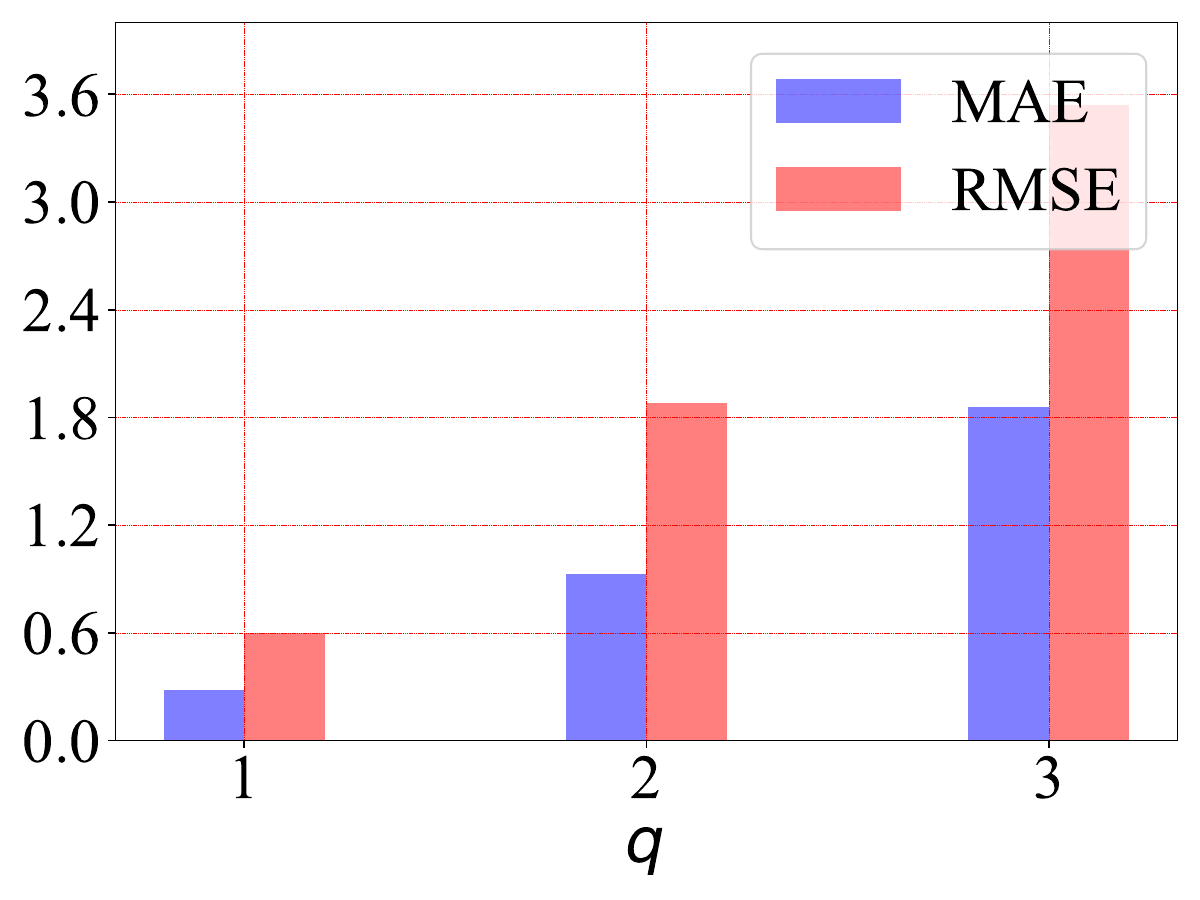} &
\includegraphics[width=1.65in]{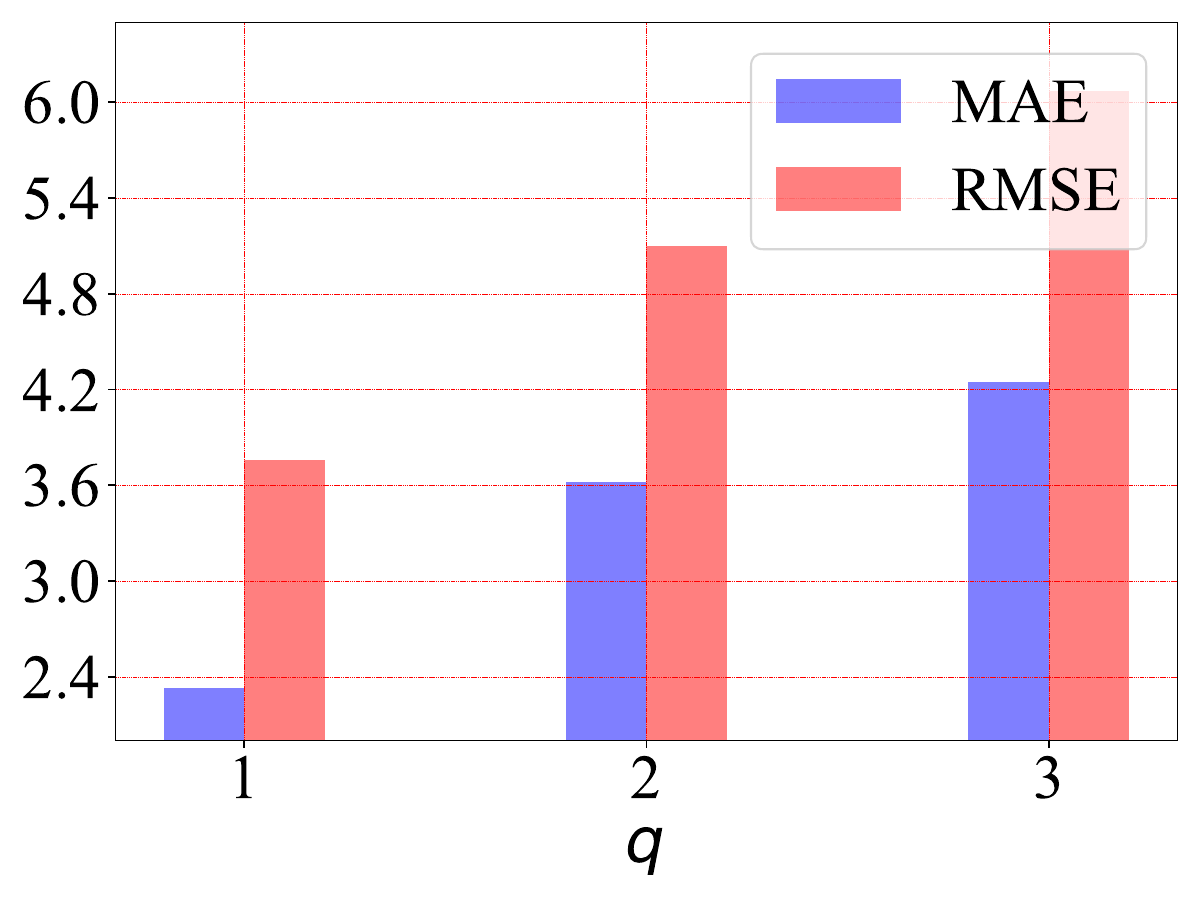} \\
(a) UBFC & (b) MMVS\\
\end{tabular}
\end{center}
\caption{Parameter variation on the number of negative STMaps when generating C-STMaps.}

\label{fig10}
\end{figure}

\subsection{Generation of C-STMaps}\label{sec6.10}
We horizontally concatenate every positive STMap and every negative STMap to create a set of C-STMaps. In this configuration, the frequency of horizontal color variation within each C-STMap changes only once. More complex C-STMaps can be constructed by concatenating every positive STMap with multiple negative STMaps. For example, assuming we have a positive STMap and two negative STMaps containing rPPG signals with frequencies of $f^a, 0.5 \times f^a, 2\times f^a$, respectively. Following the process outlined in Section~\ref{sec3.3.1}, we crop and concatenate them into a C-STMap. We then generate a text prompt to describe the relative ratios of signal frequencies across different sides of it, \ie, "the frequency of the horizontal color variation on the middle and right side is respectively 1/2 times and 2 times of that on the left side of the image". We denote the number of negative STMaps used to create C-STMaps by $q$. In Fig.~\ref{fig10}, we vary $q$ from 1 (our default setting) to 3. We can see a larger $q$ results in poorer performance because the complex frequency change in C-STMaps would confuse the encoder to accurately determine the frequency ratios between different parts.

%thereby making the model more difficult to converge.
\section{Conclusion}

This paper presents a novel {frequency-centric} self-supervised learning framework that bootstraps vision-language models (VLMs) for remote physiological measurement. It develops both generative and contrastive learning mechanisms to enhance the VLM's ability to capture the frequency of skin color temporal variation for rPPG estimation.
% digest the frequency-related knowledge in vision and text modalities.
% Our approach consists of five key stages: data augmentation, vision-text pair generation, feature extraction, STMap reconstruction and network optimization. For data augmentation, we apply spatial and frequency augmentations to generate positive and negative video samples with varying rPPG signal frequencies. We then construct their spatio-temporal maps (STMaps). For vision-text pair generation, we concatenate STMaps from positive and negative samples to create cross-sample spatio-temporal maps (C-STMaps). Then we construct text prompts to describe the relative ratios of signal frequencies across different regions within these maps. For feature extraction, we encode the visual and text embeddings using pre-trained vision and text encoders from VLM. For STMap reconstruction, we propose TMR module to reconstruct the masked image patches of C-STMaps based on linguistic clues from the text embeddings. Last, for the network optimization, we propose the vision-text contrastive loss to align vision and text modalities, the frequency contrastive loss to optimize the estimated rPPG signals across samples, the frequency ranking loss to ensure the predicted rPPG signals maintain correct frequency ranks. 
For a given facial video, our VL-phys begins by creating positive and negative spatio-temporal maps (STMaps) with varying rPPG signal frequencies. Then we horizontally concatenate them to create contrastive spatio-temporal maps (C-STMaps). We also construct text prompts to describe the relative ratios of signal frequencies across different parts of C-STMaps; this forms frequency-oriented vision-text pairs. Next, we fine-tune the pre-trained vision and text encoders of VLM using these vision-text pairs via {frequency-related} multimodal generative and contrastive tasks. These include the text-guided visual reconstruction task aimed at recovering masked image patches of C-STMaps with the guidance of text prompts, and the vision-text contrastive learning task to align vision and text modalities. Moreover, we also propose the unimodal frequency contrastive and ranking task to optimize the estimated rPPG signals across multiple video samples.
% we encode the visual and textual embeddings from these created vision-text pairs using encoders of VindLU. 
% the frequency ranking loss to ensure the predicted rPPG signals maintain correct frequency ranks.
Extensive experiments on four datasets demonstrate that our method not only outperforms existing self-supervised methods but also rivals state of the art supervised methods in HR, HRV, and RF estimations.

\section*{Declarations}
\begin{itemize}
\item Competing interests

The authors declare that they have no competing interests.
\item Data availability 

UBFC-rPPG dataset is available at https://sites.google.com/view/ybenezeth/ubfcrppg.

PURE dataset is available at https://www.tu-ilmenau.de/universitaet/fakultaeten/fakultaet-informatik-und-automatisierung/profil/institute-und-fachgebiete/institut-fuer-technische-informatik-und-ingenieurinformatik/fachgebiet-neuroinformatik-und-kognitive-robotik/data-sets-code/pulse-rate-detection-dataset-pure.

VIPL-HR dataset is available at https://vipl.ict.ac.cn/resources/databases/201811/t20181129\_32716.html.

MMVS is available at https://github.com/yuezijie/HR-estimation/tree/master.

\end{itemize}

\bibliography{sn-bibliography.bib}% common bib file

\end{document}